\documentclass[10pt,twocolumn,letterpaper]{article}

\usepackage{iccv}              
\usepackage{times}
\usepackage{epsfig}
\usepackage{graphicx}
\usepackage{amsmath}
\usepackage{amssymb}
\usepackage{times}
\usepackage{epsfig}
\usepackage{graphicx}
\usepackage{tikz}
\usepackage{xcolor}
\usepackage{amsmath}
\usepackage{amssymb}
\usepackage{xcolor}
\usepackage{booktabs}
\usepackage{textcomp}
\usepackage{xspace}
\usepackage[most]{tcolorbox}
\usepackage{caption} 
\newcommand{\NumVision}{VisNumBench\xspace}
\newcommand{\LVLMs}{MLLMs\xspace}
\definecolor{darkgray}{RGB}{169,169,169} 
\definecolor{lightgray}{RGB}{230,230,230} 

\definecolor{iccvblue}{rgb}{0.21,0.49,0.74}
\usepackage[pagebackref,breaklinks,colorlinks,allcolors=iccvblue]{hyperref}

\def\paperID{*****} 
\def\confName{ICCV}
\def\confYear{2025}

\title{\NumVision: Evaluating Number Sense of Multimodal Large Language Models}

\author{
Tengjin Weng\textsuperscript{1,2}, 
Jingyi Wang\textsuperscript{3}, 
Wenhao Jiang\textsuperscript{2}\thanks{Corresponding author.}, 
Zhong Ming\textsuperscript{1,2,4}\footnotemark[1] \\
\textsuperscript{1}College of Computer Science and Software Engineering, Shenzhen University\\
\textsuperscript{2}Guangdong Laboratory of Artificial Intelligence and Digital Economy (SZ)\\
\textsuperscript{3}Shenzhen International Graduate School, Tsinghua University\\
\textsuperscript{4}Shenzhen Technology University\\
{\tt\small wtjdsb@gmail.com, jingyi-w24@mails.tsinghua.edu.cn, cswhjiang@gmail.com, mingz@szu.edu.cn}
}

\begin{document}
\maketitle
\begin{abstract}
Can Multimodal Large Language Models (\LVLMs) develop an intuitive number sense similar to humans? Targeting this problem, we introduce Visual Number Benchmark (\NumVision) to evaluate the number sense abilities of \LVLMs across a wide range of visual numerical tasks. 
\NumVision consists of about $1,900$ multiple-choice question-answer pairs derived from both synthetic and real-world visual data, covering seven visual numerical attributes and four types of visual numerical estimation tasks.
Our experiments on \NumVision led to the following key findings:
(i) The 17 \LVLMs we tested—including open-source models such as Qwen2.5-VL and InternVL2.5, as well as proprietary models like GPT-4o and Gemini 2.0 Flash—perform significantly below human levels in number sense-related tasks.
(ii)  Multimodal mathematical models and multimodal chain-of-thought (CoT) models did not exhibit significant improvements in number sense abilities.
(iii) Stronger \LVLMs with larger parameter sizes and broader general abilities demonstrate modest gains in number sense abilities.
We believe \NumVision will serve as a valuable resource for the research community, encouraging further advancements in enhancing \LVLMs' number sense abilities. 
Code and dataset are available at \href{https://wwwtttjjj.github.io/VisNumBench/}{https://wwwtttjjj.github.io/VisNumBench/}.

\end{abstract}
\vspace{-3.5mm}
\vspace{-3mm}
\section{Introduction}\label{sec:introduction}
\begin{figure}[!t]
\label{fig:four}
\centerline{\includegraphics[width=\linewidth]{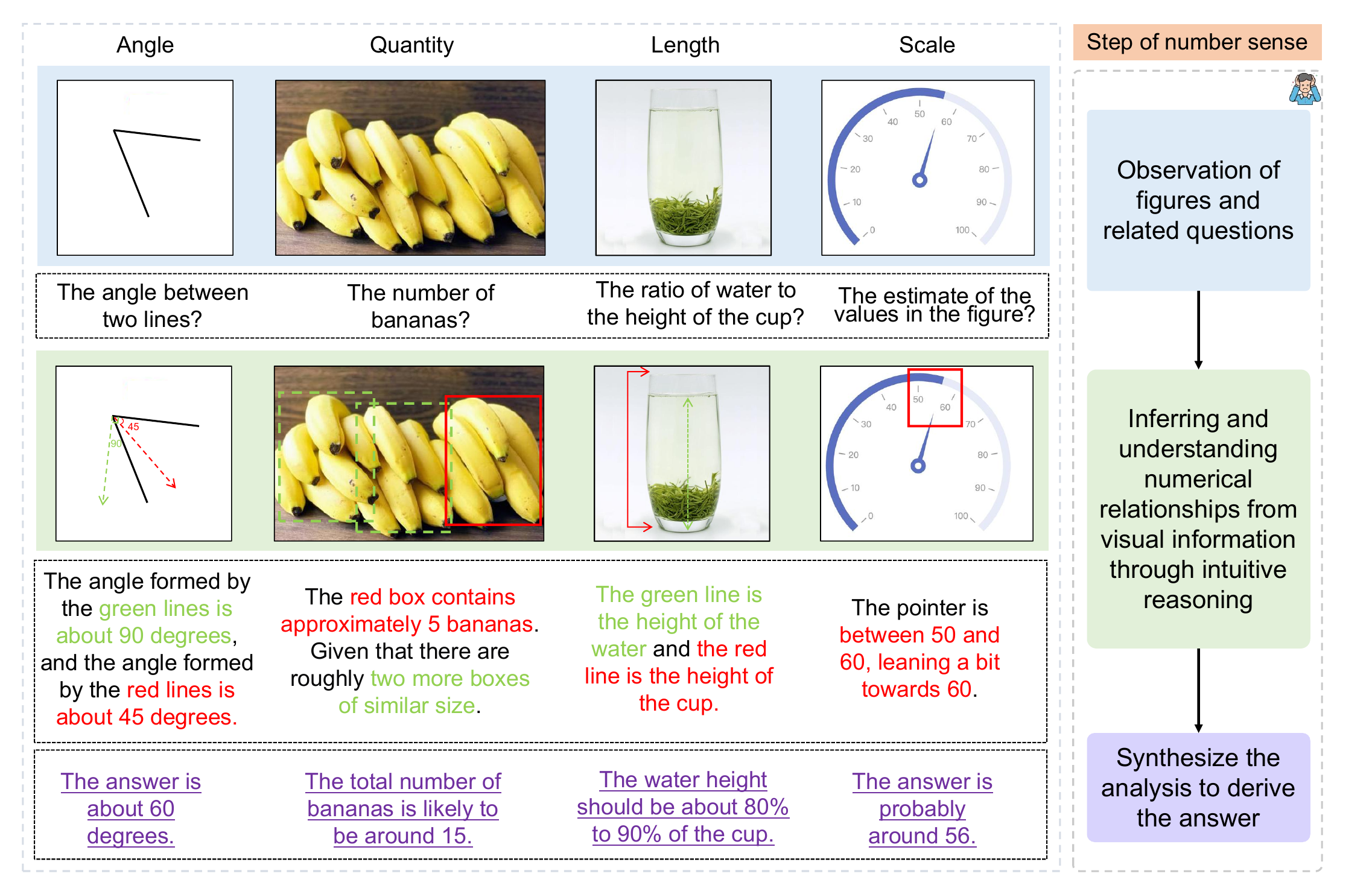}}
    \caption{
    Explanations of number sense: how humans intuitively perceive and estimate values of angle, quantity, length, and scale.
    }
    \label{fig:numbersense}
    \vspace{-4mm}
    
\end{figure}

Number sense is an innate cognitive ability shared by both humans and animals through the approximate number system~\cite{feigenson2004core}. 
It enables individuals to perceive, process, and manipulate numerical information intuitively. By fostering a deeper understanding of abstract number concepts, it facilitates the grasp of complex mathematical theories and their practical application in real-world scenarios.
Figure~\ref{fig:numbersense} illustrates the human ability to perceive and estimate numerical quantities. For instance, a person can quickly recognize a group of five bananas and intuitively estimate that there are about two more groups of the same size. By leveraging this innate number sense and grouping strategy, one can infer that the total number of bananas is approximately fifteen.

\begin{figure*}[!t]
\label{fig:overall}

\centerline{\includegraphics[width=\linewidth]{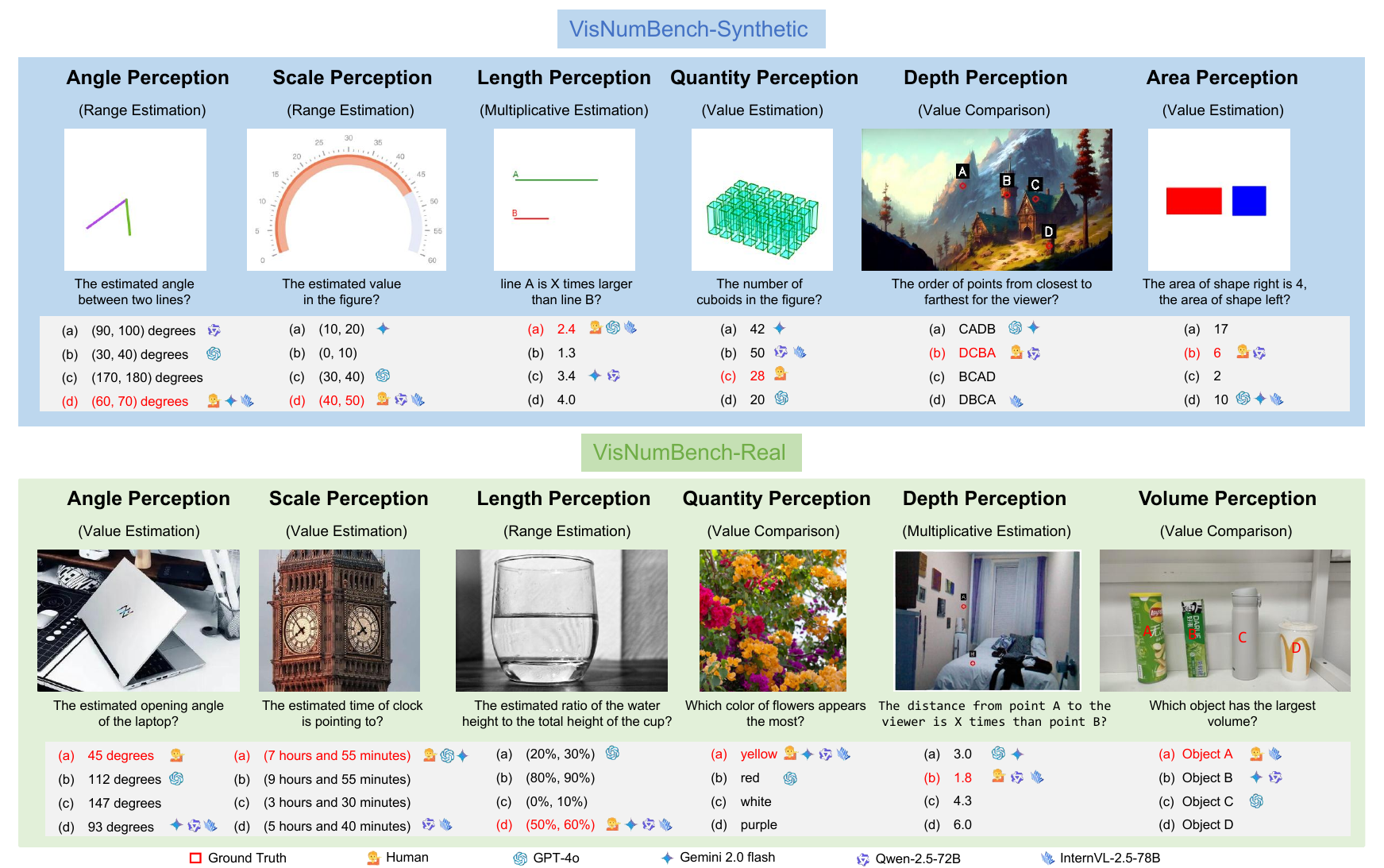}}
    \caption{
    Examples from \NumVision and responses from \LVLMs. \NumVision is divided into two subsets: \NumVision-Synthetic and \NumVision-Real. It focuses on seven key { visual numerical attributes: angle, scale, length, quantity, depth, area, and volume}. Even state-of-the-art \LVLMs often struggle to answer the questions in \NumVision accurately.
    }
    \label{fig:overall}
\end{figure*}

\begin{figure}[!t]
\label{fig:four}
\centerline{\includegraphics[width=\linewidth]{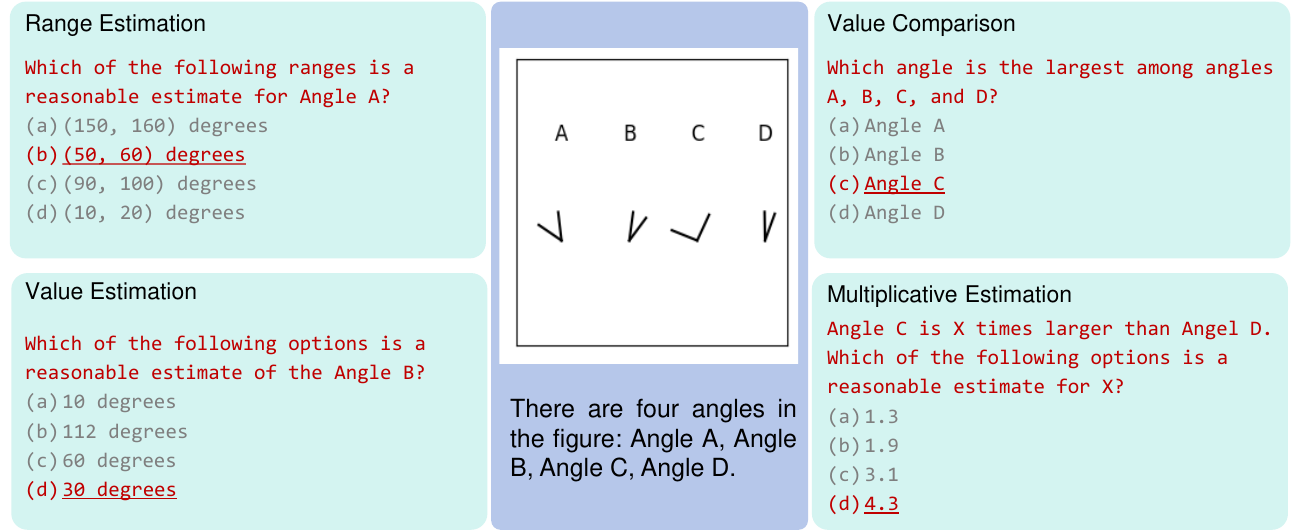}}
    \caption{
    Illustrations of four distinct visual numerical estimation tasks: range estimation, value comparison, value estimation,  and multiplicative estimation.
    }
\label{fig:four}
\vspace{-6mm}
\end{figure}

Multimodal Large Language Models (\LVLMs) have made remarkable strides in tackling complex multimodal tasks~\cite{alayrac2022flamingo,chen2024far,li2023blip,liu2024improved}. Recent research has focused on enhancing their mathematical and scientific reasoning capabilities by incorporating external tools~\cite{lu2023chameleon,wang2023scibench}. To assess these abilities, numerous benchmarks~\cite{lu2021inter,dahlgren2022clevr,masry2022chartqa,lu2023mathvista,kamoi2024visonlyqa, hu2025emobench} have been developed to evaluate the performance of \LVLMs on mathematical reasoning and numerical interpretation tasks.  
While existing benchmarks effectively assess structured numerical reasoning problems, they primarily emphasize abstract symbolic computation, mathematical problem-solving, or interpreting numerical data in textual contexts.
However, these evaluations overlook a critical aspect of human-like numerical cognition: intuitive number sense. Unlike humans, who effortlessly estimate quantities, perceive proportions, and grasp numerical relationships at a glance, \LVLMs often depend on explicit reasoning steps rather than perceptual intuition. This limitation raises fundamental questions about whether current models genuinely comprehend numerical concepts or merely manipulate them based on learned patterns in text and images.
\begin{figure}[!t]
\label{fig:randar}
\centerline{\includegraphics[width=\linewidth]{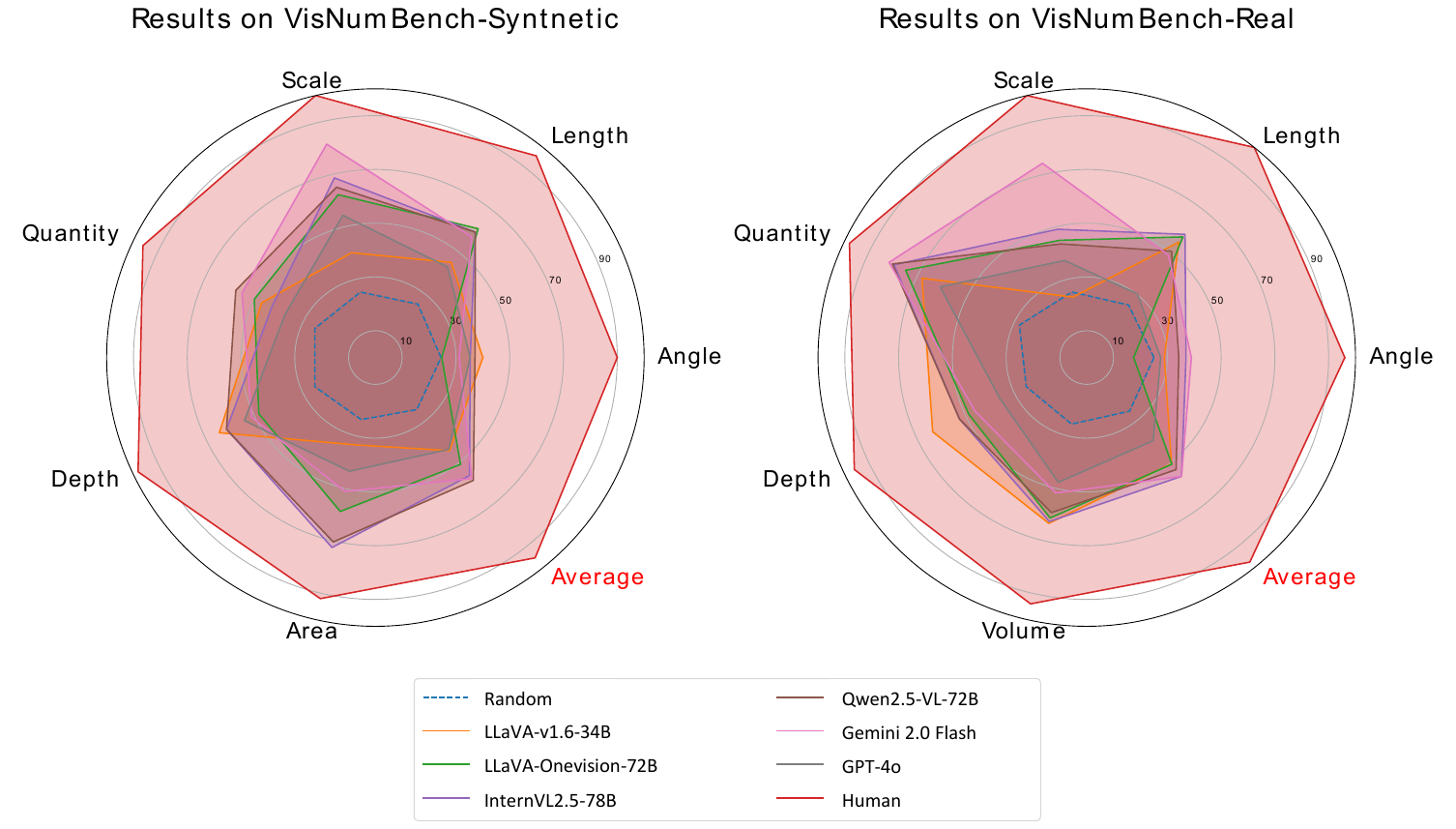}}
    \caption{
    Evaluation results of MLLMs on the VisNumBench.
    The performance of \LVLMs on \NumVision is significantly poor in terms of accuracy, whereas human performance is nearly perfect.
    }
    \label{fig:randar}
    \vspace{-6mm}
\end{figure}

\begin{table*}[!t]
\centering
    \renewcommand\arraystretch{1.2}
    \caption{
    Dataset statistics of \NumVision based on various visual numerical attributes.
    }
    \label{tab_data}
    \large
\scalebox{0.75}{
\begin{tabular}{ccccccccc}
\toprule[1pt]
VisNumBench            & Angle       & Length      & Scale     & Quantity    & Depth     & Area        & Volume      & Total         \\ \midrule
VisNumBench-Synthetic & 170         & 181         & 140       & 196         & 135       & 189         & -           & 1011          \\
VisNumBench-Real      & 149         & 162         & 143       & 147         & 154       & -           & 147         & 902           \\ \midrule
Answer Format        & 4/5 options & 3/4 options & 4 options & 3/4 options & 4 options & 4/5 options & 3/4 options & 3/4/5 options \\ \bottomrule[1pt]
\end{tabular}}
\end{table*}

In this work, we introduce the Visual Number Benchmark (\NumVision), inspired by human number sense abilities. As illustrated in Figure~\ref{fig:overall}, \NumVision is structured into two components based on different visual scenarios: \NumVision-Synthetic and \NumVision-Real. \NumVision-Synthetic comprises controlled, synthetic images in which numerical relationships are explicitly defined. \NumVision-Real contains real-world images, providing a more complex and less controlled environment. \NumVision targets seven key dimensions of visual numerical attributes through four distinct types of visual numerical estimation tasks, as depicted in Figure~\ref{fig:four}.

We evaluated 17 \LVLMs on \NumVision and found that even state-of-the-art models perform poorly on our proposed benchmark, which is shown in Figure~\ref{fig:randar}.
Furthermore, our experiments reveal that adopting multimodal mathematical models and multimodal Chain-of-Thought (CoT) models did not lead to substantial performance improvements. 
However, the performance of the latest models is better than the previous models from the same family. For example, Qwen2.5VL~\cite{qwen2.5-VL} performs better than Qwen2VL~\cite{qwen2-vl}. It seems that optimization on data, training techniques, and model architecture will help models improve their number sense ability.
In this work, we aim to advance \LVLMs toward higher levels of intelligence by developing models that enhance visual number sense abilities. 
\noindent The main contributions of this paper are listed as follows:
\begin{enumerate}  
    \item We introduce \NumVision, a comprehensive benchmark integrating diverse data sources and an automated evaluation framework to assess the numerical sense abilities of \LVLMs across various visual numerical tasks.
    \item We conduct a comprehensive evaluation of various \LVLMs on \NumVision, finding that even the most advanced models still exhibit limited numerical sense.
    \item Further experiments on historical models from the same family show that their numerical sense abilities have improved over time. To enhance this ability within a short period, more specialized optimizations in data, training techniques, and model architecture may be required.
\end{enumerate}

\section{Related Work}
\label{sec: Related Works}
\subsection{Multimodal Large Language Models}  
Recent advancements in \LVLMs have demonstrated exceptional capabilities across a wide range of tasks. Leveraging multimodal pre-training, \LVLMs have achieved outstanding performance in both open-source models~\cite{xue2024xgen, abdin2024phi, wang2024qwen2, chen2025januspro, chen2024expanding} and proprietary models~\cite{openai2023gpt4v, pichai2024our, anthropic2024claude}. Consequently, these models have been widely adopted in various domains, including mathematical reasoning~\cite{zhang2024mavis}, chart understanding~\cite{masry2023unichart}, medical image analysis~\cite{li2024llava}, and text-rich image comprehension~\cite{zhang2023llavar}. Their growing success has spurred the development of an increasing number of benchmarks to assess performance across diverse visual and linguistic tasks.
\begin{figure*}[!t]
\label{fig:source}

\centerline{\includegraphics[width=\linewidth]{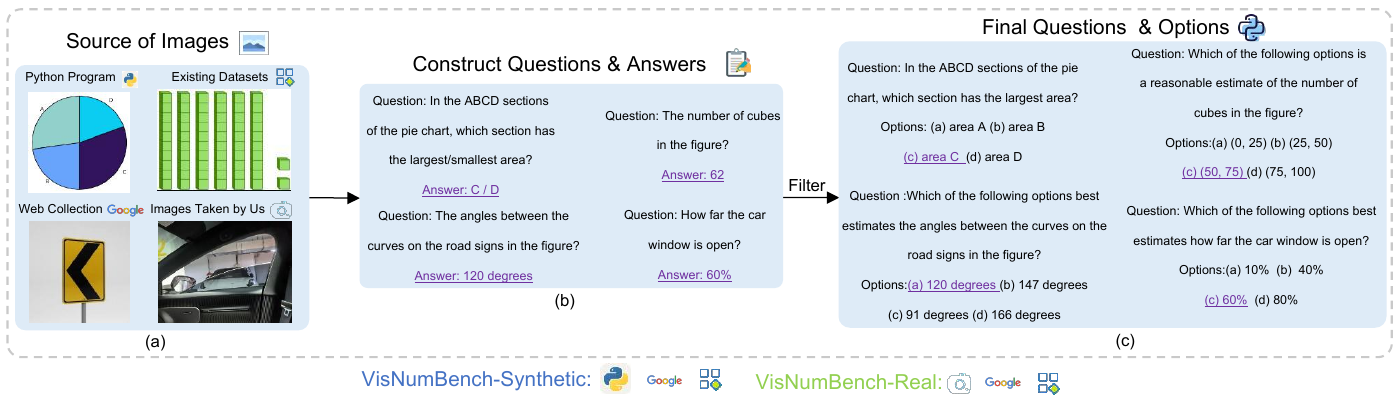}}
\vspace{-4mm}
    \caption{An illustration of the construction steps for images and questions in \NumVision-Synthetic and \NumVision-Real.}
    \label{fig:source}
    \vspace{-4mm}
\end{figure*}

\subsection{Benchmarks for \LVLMs}
Advancements in \LVLMs have led to the development of numerous benchmarks aimed at evaluating model performance across a broad spectrum of general multimodal tasks. Several recent studies~\cite{fu2023mme, yu2023mmvet, liu2023mmbench, li2023seed, li2023seed2, yang2024thinking, fu2025blink, kamoi2024visonlyqa} have introduced more comprehensive reasoning and perception multimodal benchmarks that provide extensive and holistic assessments. Beyond general multimodal tasks, specialized benchmarks have been created to evaluate the mathematical reasoning capabilities of \LVLMs. Tasks such as visual reasoning with numbers, arithmetic problem-solving, and algebraic manipulation play a crucial role in assessing both the numerical proficiency and higher-order cognitive abilities of \LVLMs. Prominent benchmarks, including MATHVISTA~\cite{lu2023mathvista}, Math-Vision~\cite{wang2024measuring}, MathOdyssey~\cite{fang2024mathodyssey}, and SMART-840~\cite{cherian2024evaluating}, are designed to test models on a diverse range of mathematical challenges, such as word problems, equation-solving, and complex multi-step reasoning.

These benchmarks aim to assess the ability of models to understand and process mathematical content in both images and text, as well as their ability to apply mathematical operations in reasoning contexts. However, evaluating \LVLMs is crucial not only for measuring their proficiency in traditional mathematical reasoning but also for understanding their ability to handle more tangible, real-world dependent mathematical perception tasks. Humans typically acquire basic mathematical knowledge through intuitive number sense, which they then apply to real-world scenarios. This intuitive understanding of numbers and their relationships is a fundamental aspect of human cognition, enabling the seamless application of mathematical concepts in everyday life.
While \LVLMs can process complex mathematical problems, they may struggle with tasks that require this intuitive number sense.
Therefore, existing benchmarks should not only evaluate the proficiency of models in traditional mathematical reasoning but also assess their ability to apply mathematical concepts in real-world contexts. This would help bridge the gap between abstract problem-solving and practical application.
\subsection{Number Sense of \LVLMs}
In the context of \LVLMs, previous research~\cite{li2022ordinalclip, wang2023learning, du2024teach} has focused on tasks that rely on ordinal regression to evaluate number sense. Examples of such tasks include Age Estimation~\cite{ricanek2006morph}, Historical Image Dating~\cite{palermo2012dating}, and Image Aesthetics Assessment~\cite{schifanella2015image}. These studies typically focus on estimating or ranking numerical attributes based on visual input, such as predicting the age of a person in an image or determining the historical context of a photograph. While these tasks assess the ability of the model to interpret numerical cues and sequences, they often focus on specific domains and may overlook the broader, more generalizable concept of number sense.
\par
Assessing the number sense of \LVLMs refers to evaluating their number sense abilities in a variety of scenarios. This includes tasks such as interpreting measurements, performing approximate calculations, comparing quantities, and identifying numerical relationships in different contexts. Such evaluations provide a better understanding of how models perform on tasks that require flexible and contextual reasoning about numbers while also enhancing their broad applicability and intelligence.

\section{\NumVision}
\label{sec:dataset}

\subsection{Overview}
We introduce \NumVision, a benchmark specifically designed to directly evaluate the intuitive numerical abilities of \LVLMs. Each instance in \NumVision comprises a figure, a multiple-choice question, and a corresponding answer label.
The dataset statistics of \NumVision are presented in Table~\ref{tab_data}.

Compared with previous benchmarks, \NumVision has the following novel features:
\begin{itemize}
\item \textbf{Comprehensive Scenario Integration:} \NumVision incorporates both controlled synthetic figures and intricate real-world scenes, enabling a thorough evaluation of the number sense abilities of \LVLMs. 

\item \textbf{Multidimensional Visual Numerical Attributes:} \NumVision encompasses seven fundamental aspects of number sense—angle, length, scale, quantity, depth, area, and volume—ensuring a rigorous and comprehensive evaluation of the numerical capabilities of \LVLMs.
\item \textbf{Comprehensive Visual Numerical Estimation Tasks:} \NumVision encompasses four distinct modes of visual numerical estimation—value comparison, value estimation, range estimation, and multiplicative estimation. These diverse tasks enable a thorough evaluation of \LVLMs' ability to estimate numerical values across different visual numerical categories.
\end{itemize}
\begin{table*}[!t]
\centering
    \renewcommand\arraystretch{1.0}
    \caption{Accuracies of MLLMs on the \NumVision-Synthetic (\%) dataset. \colorbox{darkgray}{Dark gray} and \colorbox{lightgray}{light gray}  indicate the best and the second best results among all models, respectively.
    }
    \label{tabel_synthetic}
    \large
\scalebox{0.85}{
\begin{tabular}{cccccccc}
\toprule[1pt]
\multicolumn{1}{c}{}                    & \multicolumn{1}{c}{Angle} & \multicolumn{1}{c}{Length} & \multicolumn{1}{c}{Scale}  & \multicolumn{1}{c}{Quantity} & \multicolumn{1}{c}{Depth} & \multicolumn{1}{c}{Area}  & Average   \\ \midrule
\multicolumn{1}{c}{Random}              & \multicolumn{1}{c}{24.44} & \multicolumn{1}{c}{25.41}  & \multicolumn{1}{c}{25.00}  & \multicolumn{1}{c}{25.00}    & \multicolumn{1}{c}{25.00} & \multicolumn{1}{c}{23.68} & 24.76 \\ \midrule
\multicolumn{8}{c}{{Open-source \LVLMs}}                                                                                                                                                                                              \\ \midrule
\multicolumn{1}{c}{Phi-3.5-vision}      & \multicolumn{1}{c}{19.41} & \multicolumn{1}{c}{40.88}  & \multicolumn{1}{c}{41.43}  & \multicolumn{1}{c}{26.53}    & \multicolumn{1}{c}{26.67} & \multicolumn{1}{c}{39.15} & 32.34 \\ 
\multicolumn{1}{c}{LLaVA-v1.5-7B}       & \multicolumn{1}{c}{31.18} & \multicolumn{1}{c}{30.39}  & \multicolumn{1}{c}{34.29}  & \multicolumn{1}{c}{33.16}    & \multicolumn{1}{c}{26.67} & \multicolumn{1}{c}{21.16} & 29.38 \\ 
\multicolumn{1}{c}{LLaVA-v1.5-13B}      & \multicolumn{1}{c}{{35.88}} & \multicolumn{1}{c}{30.94}  & \multicolumn{1}{c}{32.14}  & \multicolumn{1}{c}{36.73}    & \multicolumn{1}{c}{33.33} & \multicolumn{1}{c}{24.34} & 32.15 \\ 
\multicolumn{1}{c}{LLaVA-v1.6-34B}      & \multicolumn{1}{c}{\colorbox{darkgray}{40.00}} & \multicolumn{1}{c}{45.30}  & \multicolumn{1}{c}{40.00}  & \multicolumn{1}{c}{46.94}    & \multicolumn{1}{c}{\colorbox{darkgray}{64.44}} & \multicolumn{1}{c}{33.33} & 44.31 \\ 
\multicolumn{1}{c}{LLaVA-Onevision-7B}  & \multicolumn{1}{c}{25.88} & \multicolumn{1}{c}{51.38}  & \multicolumn{1}{c}{42.86}  & \multicolumn{1}{c}{38.78}    & \multicolumn{1}{c}{34.81} & \multicolumn{1}{c}{44.44} & 39.96 \\ 
\multicolumn{1}{c}{LLaVA-Onevision-72B} & \multicolumn{1}{c}{24.71} & \multicolumn{1}{c}{\colorbox{darkgray}{61.33}}  & \multicolumn{1}{c}{62.14}  & \multicolumn{1}{c}{50.00}    & \multicolumn{1}{c}{48.15} & \multicolumn{1}{c}{{58.73}} & 50.84 \\ 
\multicolumn{1}{c}{InternVL2.5-8B}      & \multicolumn{1}{c}{26.47} & \multicolumn{1}{c}{41.99}  & \multicolumn{1}{c}{49.29}  & \multicolumn{1}{c}{34.69}    & \multicolumn{1}{c}{41.48} & \multicolumn{1}{c}{46.03} & 39.66 \\ 
\multicolumn{1}{c}{InternVL2.5-38B}     & \multicolumn{1}{c}{\colorbox{lightgray}{39.41}} & \multicolumn{1}{c}{\colorbox{lightgray}{59.67}}  & \multicolumn{1}{c}{59.29}  & \multicolumn{1}{c}{54.08}    & \multicolumn{1}{c}{60.74} & \multicolumn{1}{c}{61.38} & 55.59 \\ 
\multicolumn{1}{c}{InternVL2.5-78B}     & \multicolumn{1}{c}{35.29} & \multicolumn{1}{c}{\colorbox{lightgray}{59.67}}  & \multicolumn{1}{c}{\colorbox{lightgray}{68.57}}  & \multicolumn{1}{c}{42.86}    & \multicolumn{1}{c}{\colorbox{lightgray}{61.48}} & \multicolumn{1}{c}{\colorbox{darkgray}{72.49}} & 56.18 \\ 
\multicolumn{1}{c}{Janus-Pro-7B}        & \multicolumn{1}{c}{{31.76}} & \multicolumn{1}{c}{{43.65}}  & \multicolumn{1}{c}{45.71}  & \multicolumn{1}{c}{35.71}    & \multicolumn{1}{c}{33.33} & \multicolumn{1}{c}{36.51} & 37.69 \\ 
\multicolumn{1}{c}{Qwen2.5-VL-3B}       & \multicolumn{1}{c}{30.00} & \multicolumn{1}{c}{49.17}  & \multicolumn{1}{c}{50.71}  & \multicolumn{1}{c}{32.14}    & \multicolumn{1}{c}{42.22} & \multicolumn{1}{c}{51.85} & 42.43 \\ 
\multicolumn{1}{c}{Qwen2.5-VL-7B}       & \multicolumn{1}{c}{23.53} & \multicolumn{1}{c}{53.59}  & \multicolumn{1}{c}{55.00}  & \multicolumn{1}{c}{39.29}    & \multicolumn{1}{c}{48.89} & \multicolumn{1}{c}{58.20} & 46.19 \\ 
\multicolumn{1}{c}{Qwen2.5-VL-72B}      & \multicolumn{1}{c}{37.06} & \multicolumn{1}{c}{\colorbox{lightgray}{59.67}}  & \multicolumn{1}{c}{{65.00}}  & \multicolumn{1}{c}{\colorbox{darkgray}{57.65}}    & \multicolumn{1}{c}{\colorbox{lightgray}{61.48}} & \multicolumn{1}{c}{{70.37}} & \colorbox{darkgray}{58.46} \\ \midrule
\multicolumn{8}{c}{{API-based models}}                                                                                                                                                                                               \\ \midrule
\multicolumn{1}{c}{GPT-4o}              & \multicolumn{1}{c}{35.29}      & \multicolumn{1}{c}{43.09}       & \multicolumn{1}{c}{54.29}       & \multicolumn{1}{c}{37.24}         & \multicolumn{1}{c}{54.07}      & \multicolumn{1}{c}{43.39}      &  43.72     \\ 
\multicolumn{1}{c}{Gemini 1.5 Flash}    & \multicolumn{1}{c}{26.47}      & \multicolumn{1}{c}{47.41}       & \multicolumn{1}{c}{44.40}       & \multicolumn{1}{c}{26.02}         & \multicolumn{1}{c}{23.70}      & \multicolumn{1}{c}{41.27}      & 33.33      \\ 
\multicolumn{1}{c}{Gemini 2.0 Flash}    & \multicolumn{1}{c}{31.18}      & \multicolumn{1}{c}{57.46}       & \multicolumn{1}{c}{\colorbox{darkgray}{81.43}}       & \multicolumn{1}{c}{\colorbox{lightgray}{55.10}}         & \multicolumn{1}{c}{51.11}      & \multicolumn{1}{c}{\colorbox{lightgray}{70.90}}      & \colorbox{lightgray}{57.57}      \\ 
\multicolumn{1}{c}{Gemini 1.5 Pro}      & \multicolumn{1}{c}{34.12}      & \multicolumn{1}{c}{39.23}       & \multicolumn{1}{c}{47.14}       & \multicolumn{1}{c}{40.82}         & \multicolumn{1}{c}{58.52}      & \multicolumn{1}{c}{48.15}     & 44.02      \\ \midrule 
\multicolumn{1}{c}{Human}               & \multicolumn{1}{c}{90.00} & \multicolumn{1}{c}{96.00}  & \multicolumn{1}{c}{100.00} & \multicolumn{1}{c}{96.00}    & \multicolumn{1}{c}{98.00} & \multicolumn{1}{c}{92.00} & 95.33 \\ \bottomrule[1pt]
\end{tabular}}
\end{table*}


\begin{table*}[!t]
\centering
    \renewcommand\arraystretch{1.0}
    \caption{Accuracies of MLLMs on the \NumVision-Real (\%) dataset. \colorbox{darkgray}{Dark gray} and \colorbox{lightgray}{Light gray} indicate the best and second-best results among all models, respectively.}
    \label{tabel_real}
    \large
\scalebox{0.85}{
\begin{tabular}{cccccccc}
\toprule[1pt]
\multicolumn{1}{c}{}                 & \multicolumn{1}{c}{Angle} & \multicolumn{1}{c}{Length} & \multicolumn{1}{c}{Scale}  & \multicolumn{1}{c}{Quantity} & \multicolumn{1}{c}{Depth} & \multicolumn{1}{c}{Volume} & Average   \\ \midrule
\multicolumn{1}{c}{Random}              & \multicolumn{1}{c}{25.00} & \multicolumn{1}{c}{25.00}  & \multicolumn{1}{c}{25.00}  & \multicolumn{1}{c}{27.83}    & \multicolumn{1}{c}{25.00} & \multicolumn{1}{c}{25.40}  & 25.54 \\  \midrule
\multicolumn{8}{c}{ {Open-source \LVLMs}}                                                                                                                                                                                              \\  \midrule
 \multicolumn{1}{c}{Phi-3.5-vision}      & \multicolumn{1}{c}{30.20} & \multicolumn{1}{c}{37.65}  & \multicolumn{1}{c}{27.97}  & \multicolumn{1}{c}{48.30}    & \multicolumn{1}{c}{48.70} & \multicolumn{1}{c}{29.93}  & 37.25 \\ 
\multicolumn{1}{c}{LLaVA-v1.5-7B}       & \multicolumn{1}{c}{22.82} & \multicolumn{1}{c}{32.72}  & \multicolumn{1}{c}{25.87}  & \multicolumn{1}{c}{36.73}    & \multicolumn{1}{c}{25.32} & \multicolumn{1}{c}{27.21}  & 28.49 \\ 
\multicolumn{1}{c}{LLaVA-v1.5-13B}      & \multicolumn{1}{c}{28.86} & \multicolumn{1}{c}{43.21}  & \multicolumn{1}{c}{29.37}  & \multicolumn{1}{c}{46.94}    & \multicolumn{1}{c}{49.35} & \multicolumn{1}{c}{41.50}  & 40.02 \\ 
 \multicolumn{1}{c}{LLaVA-v1.6-34B}      & \multicolumn{1}{c}{28.86} & \multicolumn{1}{c}{54.94}  & \multicolumn{1}{c}{23.08}  & \multicolumn{1}{c}{68.03}    & \multicolumn{1}{c}{\colorbox{lightgray}{63.64}} & \multicolumn{1}{c}{\colorbox{darkgray}{63.27}}  & 50.55 \\ 
\multicolumn{1}{c}{LLaVA-Onevision-7B}  & \multicolumn{1}{c}{18.12} & \multicolumn{1}{c}{44.44}  & \multicolumn{1}{c}{20.28}  & \multicolumn{1}{c}{64.63}    & \multicolumn{1}{c}{44.81} & \multicolumn{1}{c}{50.34}  & 40.58 \\ 
 \multicolumn{1}{c}{LLaVA-Onevision-72B} & \multicolumn{1}{c}{17.45} & \multicolumn{1}{c}{\colorbox{lightgray}{57.41}}  & \multicolumn{1}{c}{44.76}  & \multicolumn{1}{c}{74.83}    & \multicolumn{1}{c}{48.70} & \multicolumn{1}{c}{61.22}  & 50.78 \\ 

\multicolumn{1}{c}{InternVL2.5-8B}      & \multicolumn{1}{c}{28.86} & \multicolumn{1}{c}{34.57}  & \multicolumn{1}{c}{15.38}  & \multicolumn{1}{c}{64.63}    & \multicolumn{1}{c}{49.35} & \multicolumn{1}{c}{47.62}  & 40.13 \\ 
\multicolumn{1}{c}{InternVL2.5-38B}     & \multicolumn{1}{c}{30.20} & \multicolumn{1}{c}{51.85}  & \multicolumn{1}{c}{26.57}  & \multicolumn{1}{c}{\colorbox{darkgray}{83.67}}    & \multicolumn{1}{c}{{61.04}} & \multicolumn{1}{c}{58.50}  & 52.11 \\ 
 \multicolumn{1}{c}{InternVL2.5-78B}     & \multicolumn{1}{c}{\colorbox{lightgray}{36.91}} & \multicolumn{1}{c}{\colorbox{darkgray}{58.64}}  & \multicolumn{1}{c}{\colorbox{lightgray}{48.95}}  & \multicolumn{1}{c}{79.59}    & \multicolumn{1}{c}{52.60} & \multicolumn{1}{c}{\colorbox{lightgray}{62.59}}  & \colorbox{darkgray}{{56.54}} \\ 
 \multicolumn{1}{c}{Janus-Pro-7B}        & \multicolumn{1}{c}{22.82} & \multicolumn{1}{c}{32.10}  & \multicolumn{1}{c}{35.66}  & \multicolumn{1}{c}{48.98}    & \multicolumn{1}{c}{35.71} & \multicolumn{1}{c}{30.61}  & 34.26 \\ 
\multicolumn{1}{c}{Qwen2.5-VL-3B}       & \multicolumn{1}{c}{30.20} & \multicolumn{1}{c}{44.44}  & \multicolumn{1}{c}{35.66}  & \multicolumn{1}{c}{51.70}    & \multicolumn{1}{c}{43.51} & \multicolumn{1}{c}{49.66}  & 42.57 \\ 
\multicolumn{1}{c}{Qwen2.5-VL-7B}       & \multicolumn{1}{c}{24.16} & \multicolumn{1}{c}{38.89}  & \multicolumn{1}{c}{32.17}  & \multicolumn{1}{c}{59.18}    & \multicolumn{1}{c}{48.70} & \multicolumn{1}{c}{42.86}  & 41.02 \\ 
 \multicolumn{1}{c}{Qwen2.5-VL-72B}      & \multicolumn{1}{c}{34.23} & \multicolumn{1}{c}{50.62}  & \multicolumn{1}{c}{43.36}  & \multicolumn{1}{c}{80.27}    & \multicolumn{1}{c}{52.60} & \multicolumn{1}{c}{59.18}  & \colorbox{lightgray}{53.33} \\  \midrule
\multicolumn{8}{c}{{API-based models}}                                                                                                                                                                                                \\  \midrule
 \multicolumn{1}{c}{GPT-4o}              & \multicolumn{1}{c}{27.52}      & \multicolumn{1}{c}{30.25}       & \multicolumn{1}{c}{37.06}       & \multicolumn{1}{c}{60.54}         & \multicolumn{1}{c}{35.71}      & \multicolumn{1}{c}{47.62}       & 39.58      \\ 
 \multicolumn{1}{c}{Gemini 1.5 Flash}    & \multicolumn{1}{c}{14.77}      & \multicolumn{1}{c}{35.80}       & \multicolumn{1}{c}{26.57}       & \multicolumn{1}{c}{57.14}         & \multicolumn{1}{c}{24.68}      & \multicolumn{1}{c}{43.54}       & 33.70      \\ 
\multicolumn{1}{c}{Gemini 2.0 Flash}    & \multicolumn{1}{c}{\colorbox{darkgray}{38.93}}      & \multicolumn{1}{c}{48.77}       & \multicolumn{1}{c}{\colorbox{darkgray}{74.14}}       & \multicolumn{1}{c}{\colorbox{lightgray}{81.63}}         & \multicolumn{1}{c}{46.10}      & \multicolumn{1}{c}{51.70}       & \colorbox{darkgray}{56.54}      \\ 
\multicolumn{1}{c}{Gemini 1.5 Pro}      & \multicolumn{1}{c}{30.20}      & \multicolumn{1}{c}{45.68}       & \multicolumn{1}{c}{27.97}       & \multicolumn{1}{c}{68.03}         & \multicolumn{1}{c}{\colorbox{darkgray}{64.29}}      & \multicolumn{1}{c}{55.10}       & 48.67      \\ \midrule
\multicolumn{1}{c}{Human}               & \multicolumn{1}{c}{96.00} & \multicolumn{1}{c}{100.00} & \multicolumn{1}{c}{100.00} & \multicolumn{1}{c}{98.00}    & \multicolumn{1}{c}{96.00} & \multicolumn{1}{c}{94.00}  & 97.33 \\ \bottomrule[1pt]
\end{tabular}}
\end{table*}

\subsection{Data Collection Process}
\subsubsection{Source of Images}
The development of \NumVision required gathering figures from diverse sources, as depicted in part (a) of Figure~\ref{fig:source}.
\begin{itemize} 
\item \textbf{Python Program.} 
We developed a series of Python scripts based on Matplotlib\footnote{\url{https://matplotlib.org/}} to generate figures by randomly sampling parameters, which are also stored for future use.
This approach allows precise control over various numerical properties, ensuring a well-balanced data distribution and minimizing potential biases.
\item \textbf{Existing Datasets.} To enhance the diversity of the proposed benchmark and leverage existing high-quality data, we incorporated figures from multiple well-established datasets~\cite{lu2023mathvista, kamoi2024visonlyqa, fu2025blink, zhang2016single, silberman2012indoor, wang2020tartanair}, covering a broad range of numerical and spatial perception scenarios.
\item \textbf{Web Collection.} To incorporate more natural and diverse visual data, we collected figures with numerical information from public web sources~\cite{wallpaperscraft, google_images, echarts}. These figures were carefully curated and filtered to ensure relevance and clarity before designing the corresponding questions.

\item \textbf{Images Taken by Us.} To better reflect real-world conditions, we manually constructed various number sense scenarios and captured images using a camera. These scenes encompassed real-world measurements, physical counting tasks, and estimation challenges under natural lighting and occlusion conditions.
\end{itemize}

\subsubsection{QA Pair Construction and Quality Control}
As shown in parts (b) and (c) in Figure~\ref{fig:source}, we involve QA pairs for images from different sources. For the figures generated by the Python program, we manually design different questions based on the specific characteristics of each figure and generate corresponding annotations using the parameters saved.
We can design the question such as: ``\textit{In the ABCD sections of the pie chart, which section has the largest/smallest area?}" Based on the parameters saved, the answer would be ``\textit{C / D}". For images from other sources, we manually designed questions and annotated them based on different numerical attributes of the images. In addition, we can generate different numerical estimation tasks based on the answer type. For example, for the question: ``\textit{Which of the following options is a reasonable estimate of the number of cubes in the figure?"}, when the answer is ``\textit{(50, 75)"}, it corresponds to a range estimation task. On the contrary, if the answer is ``\textit{62"}, it belongs to the value estimation task.

We employ a combination of automated and manual methods to design distractors. For numerical answers, we generate alternative options that are easily confusable with the correct answer. Some distractors are constructed based on their inherent properties (e.g., areas A, B, and D in part (c) of Figure~\ref{fig:source}). These distractors are designed to align with human perceptual biases, appearing plausible yet distinguishable from the correct answer.
To ensure the high quality of \NumVision, we meticulously reviewed all collected data and filtered out any ambiguous or unclear entries. More details of the data construction process can be found in Appendix~\ref{data}.

\section{Experiments}
\label{sec:experiments}
We evaluate 17 well-known \LVLMs from 8 model families, including 13 open-source models: Phi-3.5-vision~\cite{abdin2024phi}, LLaVA-v1.5 (7B, 13B), and LLaVA-v1.6-34B~\cite{liu2023llava}, LLaVA-Onevision (7B, 72B)~\cite{li2024llava}, Qwen2.5-VL (3B, 7B, 72B)~\cite{qwen2.5-VL}, InternVL2.5 (8B, 38B, 78B)~\cite{chen2024expanding}, and Janus-Pro-7B~\cite{chen2025januspro}. Additionally, we assess 4 proprietary models: GPT-4o~\cite{hurst2024gpt}, Gemini 1.5 Flash, Gemini 2.0 Flash, and Gemini 1.5 Pro~\cite{pichai2024our}. 


We randomly selected $600$ samples ($50$ QA pairs from each numerical attribute), with $300$ sourced from \NumVision-Synthetic and $300$ from \NumVision-Real. Human evaluators independently answered each question and provided assessments. Accuracy (\%) is reported for all experimental results, and all the results are provided in Tables~\ref{tabel_synthetic} and \ref{tabel_real}.

\subsection{Evaluation Results and Analysis}
From Tables~\ref{tabel_synthetic} and \ref{tabel_real}, we observe that the performance of MLLMs is not comparable to that of humans.  
Among the seven types of questions, quantity-related tasks appear to be the easiest, while angle-related tasks are the most difficult.  
This is possibly because the amount of training data available for quantity-related tasks is significantly greater than that for angle-related tasks.  
By comparing the evaluations on synthetic and real images, we find that the performance of the same model does not exhibit significant variance.  
Thus, in terms of numerical reasoning ability, both synthetic and real images present similar challenges for existing MLLMs.  
Moreover, we observe that the best open-sourced model performs comparably to the best closed-source models. 
\begin{figure*}[!t]
\label{fig:heatmap}
\centerline{\includegraphics[width=0.85\linewidth]{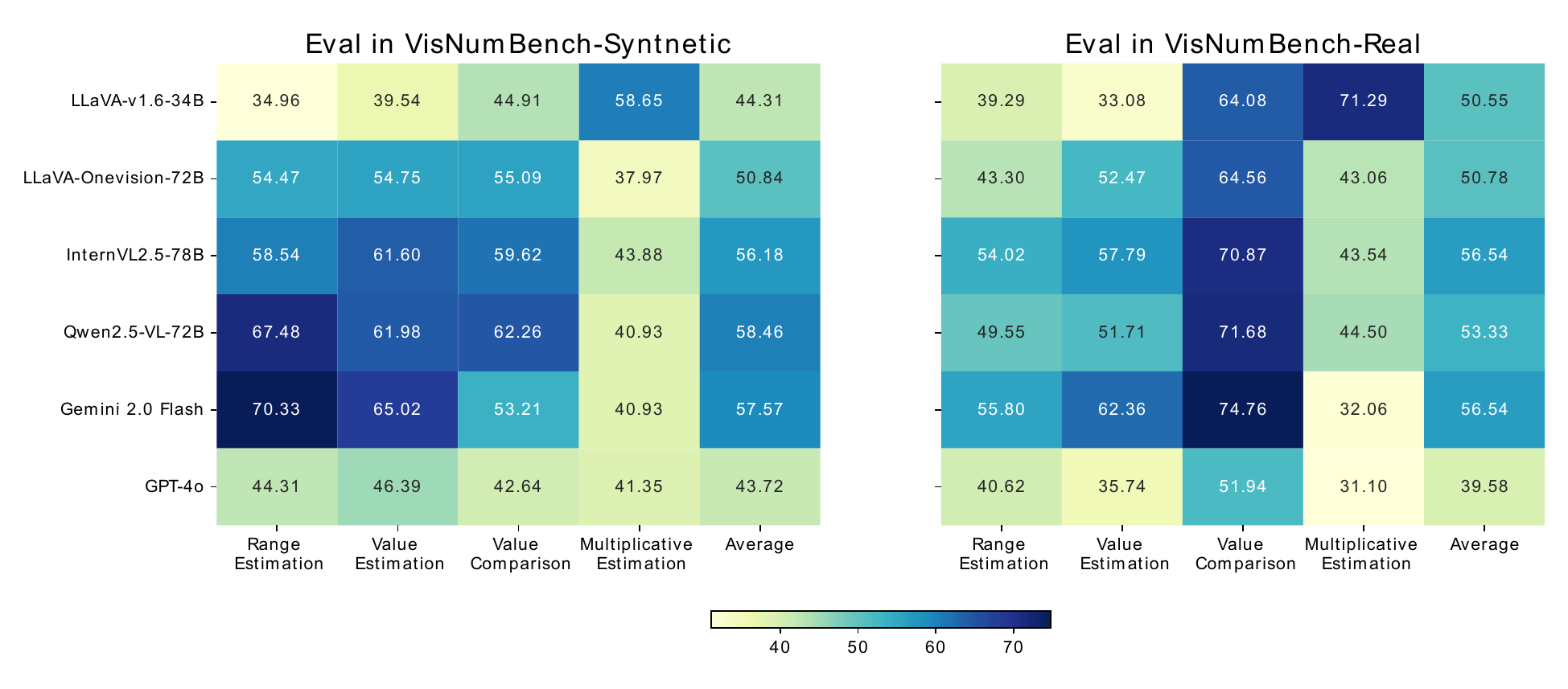}}
    \caption{Confusion matrices (\%) of \LVLMs on the \NumVision-Synthetic and \NumVision-Real datasets across different visual numerical estimation tasks.}
    \label{fig:heatmap}
    \vspace{-4mm}
\end{figure*}

More detailed analyses and discussions are provided in the following subsections.

\vspace{-1mm}
\subsubsection{Performance on \NumVision-Synthetic}
\label{sec:mainresult1}
Table~\ref{tabel_synthetic} presents the results for various \LVLMs on the \NumVision-Synthetic dataset. Among the open-source models, Qwen2.5-VL-72B achieves the best performance, with an average accuracy of $58.46\%$.
InternVL2.5-38B, InternVL2.5-78B, and LLaVA-v1.6-34B also demonstrate strong performance, each achieving either the best or the second-best accuracy in at least two tasks.
LLaVA-v1.6-34B attains the highest accuracy in angle and depth estimation; however, its overall average accuracy is only $44.31\%$.
LLaVA-Onevision-72B also performs well, achieving the highest accuracy in length estimation at $61.33\%$.
In general, models with larger parameter sizes tend to exhibit superior performance, aligning with the intuition that larger models can better capture complex numerical relationships and fine-grained visual patterns.

In the API-based models, Gemini 2.0 Flash demonstrates the best performance, achieving an average accuracy of $57.57\%$.
In contrast, GPT-4o and Gemini 1.5 Pro exhibit comparable performance, albeit with lower average accuracies.
Gemini 1.5 Flash yields the weakest performance, with an average accuracy of $33.33\%$.
Notably, certain open-source models perform on par with or even surpass proprietary models, suggesting that the disparity in numerical reasoning capabilities between open-source and closed-source models is minimal.

\subsubsection{Performance on \NumVision-Real}
\label{sec:mainresult2}
Accuracy on the \NumVision-Real dataset shows similar trends. InternVL2.5-78B and Gemini 2.0 Flash stand out with an average accuracy of $56.54\%$, achieving near-optimal results across multiple tasks, as shown in Table~\ref{tabel_real}.
InternVL2.5-38B attained an exceptionally high accuracy of $83.67\%$ on the quantity task, while LLaVA-v1.6-34B excelled in the volume task, achieving the highest scores.
Qwen2.5-VL-72B demonstrated relatively balanced performance, yielding a suboptimal average accuracy of $53.33\%$.

Surprisingly, Gemini 1.5 Pro achieved the highest accuracy in depth estimation, reaching $64.29\%$. However, its overall average accuracy remained unsatisfactory. Other proprietary models, such as GPT-4o and Gemini 1.5 Flash, exhibited relatively weaker performance. In general, the accuracy on \NumVision-Real is lower than that on \NumVision-Synthetic, likely due to the increased complexity and variability of real-world images.

\begin{figure*}[!t]
\label{fig:LNMsCOT}
\centerline{\includegraphics[width=\linewidth]{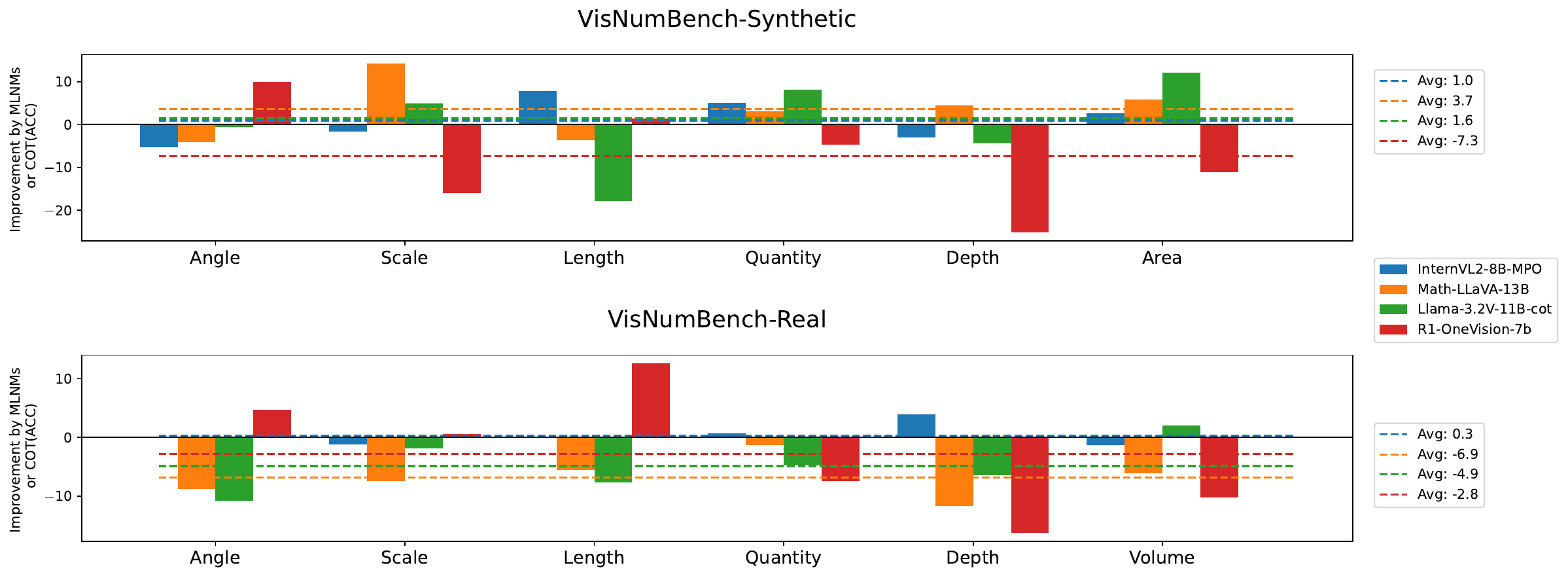}}
    \caption{
    Improvements brought by multimodal mathematical models (InternVL2-8B-MPO and Math-LLaVA-13B) and multimodal CoT models (Llama-3.2V-11B-cot and R1-OneVision-7B). Table~\ref{table:COT-S} and Table~\ref{table:COT-R} in the appendix provide detailed results.
    }
    \label{fig:LNMsCOT}
    \vspace{-2mm}
\end{figure*}

\subsubsection{Performance on Different Visual Numerical Estimation Tasks}
\label{sec:mainresult3}
As we analyze performance across different numerical estimation tasks, Figure~\ref{fig:heatmap} reveals that in the synthetic scenario, Gemini 2.0 Flash and Qwen2.5-VL-72B achieve the highest performance across all visual numerical
estimation tasks, particularly in range estimation, value estimation, and value comparison, where their accuracies consistently exceed 60\%. In contrast, GPT-4o exhibits the lowest performance in all tasks, especially in value comparison and multiplicative estimation.
In the real-world scenario, most models achieved their best performance in value comparison tasks, which are also the easiest for humans. Although Gemini 2.0 Flash and InternVL2.5-78B continue to perform well in most tasks, their performance in multiplicative estimation has declined compared to the synthetic scenario. Additionally, GPT-4o continues to perform poorly across all tasks, particularly in multiplicative estimation and value comparison, where it falls significantly behind other models.

Notably, in the multiplicative estimation task, LLaVA-v1.6-34B outperforms all other models by a significant margin. This suggests that certain models may be more specialized for specific types of tasks, and further fine-tuning could enhance performance across different tasks.

\subsection{Further Analysis}\label{sec:analysis}
\textbf{How do math-special models perform on the \NumVision?}
To investigate the number sense abilities of math-special models,  
we introduce two multimodal mathematical models: (1) InternVL2-8B-MPO~\cite{wang2024enhancing}, initialized from InternVL2-8B~\cite{chen2024internvl} and fine-tuned on the large-scale multimodal reasoning preference dataset MMPR~\cite{wang2024enhancing}, achieving an accuracy of 65.65\% on MathVista; (2) Math-LLaVA-13B~\cite{shi2024math}, initialized from LLaVA-v1.5-13B and fine-tuned on the MathV360K~\cite{shi2024math} dataset.
As shown in Figure~\ref{fig:LNMsCOT}, InternVL2-8B-MPO achieved a $1.0\%$ improvement in synthetic scenarios and a $0.3\%$ increase in real-world scenarios. Its enhancements are task-specific rather than universally effective across different number sense challenges. In contrast, Math-LLaVA-13B exhibited a polarized performance trend: while it improved by $3.7\%$ on the synthetic dataset, its accuracy declined by $6.9\%$ in real-world scenarios. This suggests that although the model benefits from training on synthetic data, it struggles to generalize to the complexity and variability of real-world number sense tasks. Relying solely on synthetic data may be insufficient to enhance number sense capabilities in real-world applications. Additional strategies, such as incorporating more diverse real-world training data or refining model architectures, may be necessary to achieve meaningful improvements.
\begin{table}[!t]
\centering
\renewcommand\arraystretch{1.2}
\caption{Comparisons of the performance of models from the Qwen-VL family and the InternVL family in synthetic and real-world scenes. Table~\ref{table:generality} in the appendix provides detailed results.}
\label{time_table}
\scalebox{0.9}{
\begin{tabular}{ccc}
\toprule[1pt]
                & \multicolumn{1}{c}{Average (Synthetic)} & \multicolumn{1}{c}{Average (Real)} \\
\midrule
Qwen2-VL-2B     & 31.85 & 24.94 \\
Qwen2.5-VL-3B   & 42.24\textcolor{green}{\((\uparrow +10.39)\)} & 42.57\textcolor{green}{\((\uparrow +17.63)\)} \\
\midrule
Qwen2-VL-7B     & 41.25 & 41.91 \\
Qwen2.5-VL-7B   & 46.19\textcolor{green}{\((\uparrow +4.94)\)} & 41.02\textcolor{red}{\((\downarrow -0.89)\)} \\
\midrule
Qwen2-VL-72B    & 54.20 & 46.56 \\
Qwen2.5-VL-72B  & 58.46\textcolor{green}{\((\uparrow +4.26)\)} & 53.33\textcolor{green}{\((\uparrow +6.77)\)} \\
\midrule
InternVL2-8B    & 39.56 & 39.58 \\
InternVL2.5-8B  & 39.66\textcolor{green}{\((\uparrow +0.10)\)} & 40.13\textcolor{green}{\((\uparrow +0.55)\)} \\
\midrule
InternVL2-40B   & 45.50 & 45.12 \\
InternVL2.5-38B & 55.59\textcolor{green}{\((\uparrow +10.09)\)} & 52.11\textcolor{green}{\((\uparrow +6.99)\)} \\
\bottomrule[1pt]
\end{tabular}}
\end{table}

\noindent \textbf{How do the multimodal reasoning models perform?}
To examine whether reasoning techniques can enhance the number sense abilities of \LVLMs, we evaluate two multimodal reasoning models: Llama-3.2V-11B-cot\footnote{\url{https://huggingface.co/Xkev/Llama-3.2V-11B-cot}}~\cite{xu2024llava} and R1-OneVision-7B\footnote{\url{https://github.com/Fancy-MLLM/R1-Onevision}}.
Llama-3.2V-11B-cot is trained using LLaVA-o1-100k~\cite{xu2024llava}, achieving a $6.2\%$ performance improvement on MathVista compared to Llama-3.2-11B-Vision-Instruct~\cite{llama3.2-vision-instruct}.
R1-OneVision-7B, trained with a rule-based reinforcement learning technique, attains an accuracy of $44.06\%$ on Mathverse~\cite{zhang2024mathverse}.
Accordingly, we assess these models on our benchmark. The results, presented in Figure~\ref{fig:LNMsCOT}, indicate that neither Llama-3.2V-11B-cot nor R1-OneVision-7B achieved the expected performance gains. On the contrary, their accuracy dropped significantly—except for a modest $1.6\%$ improvement by Llama-3.2V-11B-cot in synthetic scenarios—especially in real-world settings. These findings suggest that developing reasoning techniques specifically tailored for number sense abilities may be necessary.
\noindent \textbf{
What helps improve the performance?
}
To determine the factors contributing to the improvement of number sense ability in MLLMs, we evaluate historical models from the same family over time, specifically the Qwen-VL family and the InternVL family. The results are presented in Table~\ref{time_table}.
As observed, the performance of the latest models generally surpasses that of their predecessors.
By comparing Qwen2-VL~\cite{qwen2-vl} with Qwen2.5-VL~\cite{qwen2.5-VL}, as well as InternVL2 with InternVL2.5~\cite{internvl2.5}, we observe improvements in several aspects: (1) data scale and quality, (2) a more powerful encoder, (3) model architecture, and (4) training strategy.
These findings suggest that further exploration in these directions is essential for enhancing the number sense abilities of MLLMs.

\section{Conclusion}
\label{sec:conclusion}
In this work, we introduce \NumVision, a novel benchmark designed to evaluate \LVLMs on core number sense abilities that are inadequately addressed by existing evaluation benchmarks. Our assessment of 17 \LVLMs uncovers substantial deficiencies in their capacity to demonstrate human-like number sense.  Even the most advanced models still demonstrate limited numerical sense abilities. Further experiments on historical models from the same family show that to enhance this ability within a short period, more specialized optimizations in data, training techniques, and model architecture may be required.

\section{Acknowledge}
This research was supported by the National Natural Science Foundation of China under Grant No.~62272315.

{
    \small
    \bibliographystyle{ieeenat_fullname}
    \bibliography{custom}
}

\clearpage
\setcounter{page}{1}
\appendix

\section{Appendix Outline}
In the supplementary materials, we provide:  
\begin{itemize}  
    \item A detailed description of the VisNumBench construction process (Appendix~\ref{data});  
    \item The evaluation setup and comprehensive results for VisNumBench sub-experiments (Appendix~\ref{model});  
    \item Additional visualizations (Appendix~\ref{visual}).  
\end{itemize}

\section{Details of VisNumBench Construction}
\label{data}

 \noindent \textbf{Angle}
The task may involve recognizing angles in both 2D and 3D contexts, such as the angles between intersecting lines or the angle between the viewpoint and an object.
The figures for \NumVision-Synthetic are either generated using Python programs or sourced from VisOnlyQA~\cite{kamoi2024visonlyqa}, whereas the images for \NumVision-Real are either captured by the authors or collected from Google Images~\cite{google_images}.

 \noindent \textbf{Length}
Based on both synthetic and real-world scenes, we designed a variety of question-answering tasks, including relative length comparison, multiple segment estimation, and the estimation of object length, height, and proportion, among others.
The figures in \NumVision-Synthetic are either generated using Python programs or sourced from MathVista~\cite{lu2023mathvista}, whereas the figures in \NumVision-Real are captured by the authors or collected from Google Images~\cite{google_images}.

 \noindent \textbf{Scale}
We provide figures illustrating the coordinates of a point in a coordinate system, the time indicated on a clock, and the temperature displayed on a thermometer.  
The figures for \NumVision-Synthetic are generated by Python programs, sourced from MathVista~\cite{kamoi2024visonlyqa}, while other figures originate from Google Images~\cite{google_images} and ECharts~\cite{echarts}.  
In contrast, the statistics for \NumVision-Real are either captured by the authors or collected from Google Images~\cite{google_images}.

 \noindent \textbf{Quantity}
Each figure contains a varying number of objects, such as points or triangles in synthetic scenarios, or hot air balloons or pets in real-world scenarios.
The figures for \NumVision-Synthetic are either generated using Python programs or obtained from MathVista~\cite{lu2023mathvista}, whereas the figures for \NumVision-Real are sourced from Google Images~\cite{google_images} and the ShanghaiTech dataset~\cite{zhang2016single}.

 \noindent \textbf{Depth}
We further refine the ``Relative Depth'' task in BLINK~\cite{fu2025blink} by incorporating additional choice points and introducing new question-answer formats. \LVLMs will be presented with images containing objects at varying depths, requiring them to determine the correct depth order or estimate the relative distances between objects.  

The figures for \NumVision-Synthetic are obtained from the WallpapersCraft website~\cite{wallpaperscraft} or sourced from MathVista~\cite{lu2023mathvista} and VSLAM-TartanAir~\cite{wang2020tartanair}. The figures for \NumVision-Real are sourced from BLINK~\cite{fu2025blink} and the NYU Depth Dataset V2~\cite{silberman2012indoor}.

 \noindent \textbf{Area}
\NumVision-Synthetic contains comparisons and estimations of object area sizes for both identical and different shapes, as well as their multiplicative relationships.
The figures in \NumVision-Synthetic are either generated by Python programs or obtained from VisOnlyQA~\cite{kamoi2024visonlyqa}.
 
 \noindent \textbf{Volume}
Objects are presented from different perspectives and in various sizes, requiring \LVLMs to infer relative volume sizes and proportions based on visible dimensions and depth. The figures for \NumVision-Real are obtained through camera capture or sourced from Google Images~\cite{google_images}.

\begin{figure*}[!t]
\centerline{\includegraphics[width=0.9\linewidth]{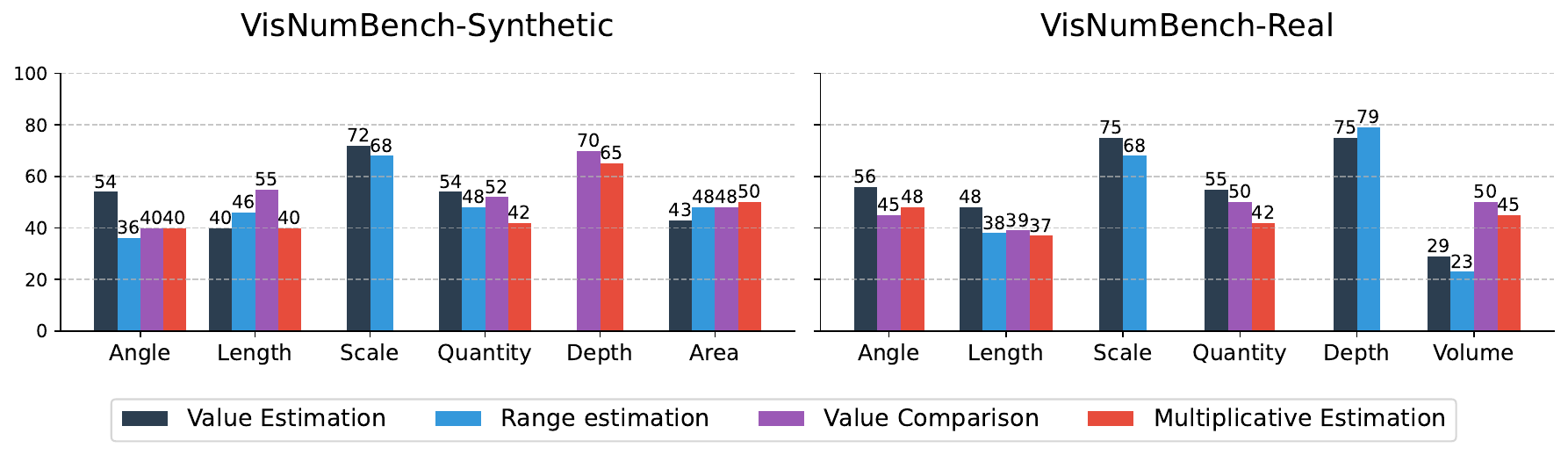}}
    \caption{Dataset statistics of \NumVision\ based on various visual numerical estimation tasks.}
    \label{fig:image_table1}
\end{figure*}

\begin{figure*}[!t]
\centerline{\includegraphics[width=\linewidth]{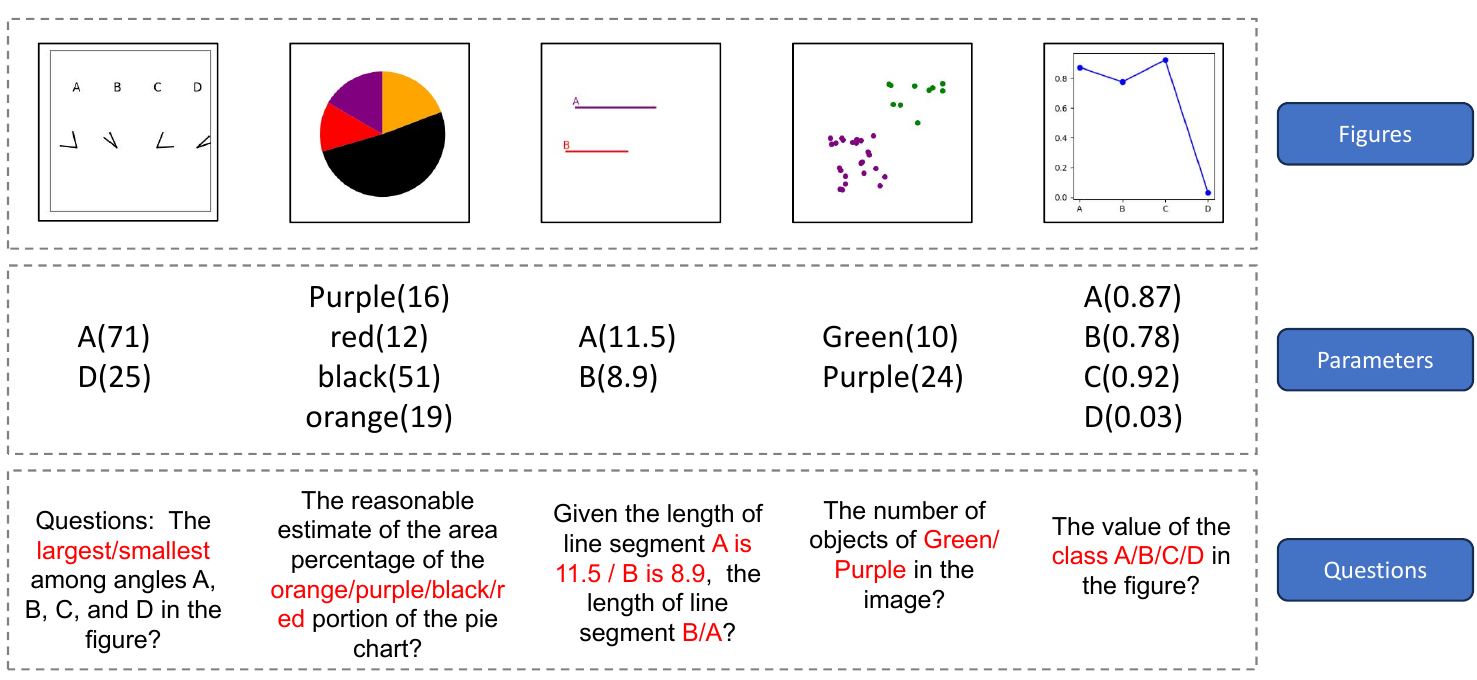}}
    \caption{Examples of the data generated by Python and manually designed questions.}
\label{table:pythondata}
\end{figure*}

More details of data construction are shown in the Tables~\ref{table:data_s}, ~\ref{table:data_r}.
Figure~\ref{fig:image_table1} shows the dataset statistics of VisNumBench based on various visual numerical estimation tasks.
Figure~\ref{table:pythondata} shows how to build a QA pair based on images generated by a Python script. Python-generated images are $500 \times 500$, and all others are resized with the longer side capped at 500 pixels.

\begin{table}[h]
\centering
    \renewcommand\arraystretch{1} 
    \caption{The source distribution of different visual numerical attributes on the  VisNumBench-Synthetic set.
    }
    \label{table:data_s}
    \large
\scalebox{0.7}{
\begin{tabular}{ccccc}
\toprule[1pt]
         & Python Program & Web Collection & Other Dataset & Total \\ \midrule
Angle    & 138            & 0              & 32            & 170   \\
Length   & 160            & 0              & 21            & 181   \\
Scale    & 77             & 50             & 13            & 140   \\
Quantity & 185            & 0              & 11            & 196   \\
Depth    & 0              & 70             & 65            & 135   \\
Area     & 139            & 0              & 50            & 189  \\ 
Total     &699             & 120              & 192            & 1011  \\ \bottomrule
\end{tabular}}
\end{table}
\begin{table}[h]
\centering
    \renewcommand\arraystretch{1}
    \caption{
     The source distribution of different visual numerical attributes on the VisNumBench-Real set.
}
    \label{table:data_r}
    \large
\scalebox{0.7}{
\begin{tabular}{ccccc}
\toprule[1pt]
         & Image Taken by Us & Web Collection & Other Dataset & Total \\ \midrule
Angle    & 91             & 58             & 0             & 149   \\
Length   & 144            & 18             & 0             & 162   \\
Scale    & 21             & 122            & 0             & 143   \\
Quantity & 0              & 113            & 34            & 147   \\
Depth    & 0              & 0              & 154           & 154   \\
Volume   & 140            & 7              & 0             & 147   \\ 
Total    & 396           & 318            & 188           & 902  \\\bottomrule
\end{tabular}}
\end{table}

\section{Evalution Details}
\label{model}
\subsection{Model Access}
This section provides details on model access and parameter settings (refer to Table~\ref{model_access}). The model responses presented in this paper were collected between January 1 and February 28, 2025. We set $\mathit{max\_new\_tokens} \geq 512$, while all other parameters were kept at their default values.

\begin{table}[h]
\centering
    \renewcommand\arraystretch{1}
    \caption{The MLLMs evaluated in this paper. This table presents the model names (Hugging Face repository name or Official API name).}
    \label{model_access}
    \large
\scalebox{0.7}{
\begin{tabular}{ll}
\toprule[1pt]
Phi-3.5-vision      & microsoft/Phi-3.5-vision-instruct        \\
LLaVA-v1.5-7B       & liuhaotian/llava-v1.5-7b                 \\
LLaVA-v1.5-13B      & liuhaotian/llava-v1.5-13b                \\
LLaVA-v1.6-34B      & liuhaotian/llava-v1.6-34b                \\
LLaVA-Onevision-7B  & llava-hf/llava-onevision-qwen2-72b-si-hf  \\
LLaVA-Onevision-72B & llava-hf/llava-onevision-qwen2-72b-ov-hf \\
InternVL2.5-8B      & OpenGVLab/InternVL2\_5-8B                \\
InternVL2.5-38B     & OpenGVLab/InternVL2\_5-38B               \\
InternVL2.5-78B     & OpenGVLab/InternVL2\_5-78B               \\
Janus-Pro-7B        & deepseek-ai/Janus-Pro-7B                 \\
Qwen2.5-VL-3B       & Qwen/Qwen2.5-VL-3B-Instruct              \\
Qwen2.5-VL-7B       & Qwen/Qwen2.5-VL-7B-Instruct              \\
Qwen2.5-VL-72B      & Qwen/Qwen2.5-VL-72B-Instruct             \\ \midrule
GPT-4o              & gpt-4o-2024-08-06                        \\
Gemini 1.5 Flash    & gemini-1.5-flash                     \\
Gemini 2.0 Flash    & gemini-2.0-flash                                 \\
Gemini 1.5 Pro      & gemini-1.5-pro-002                      \\ \bottomrule[1pt]
\end{tabular}}
\end{table}



\subsection{Detail of Evaluation}
\textbf{Prompt}. Table~\ref{prompt} shows the prompts for evaluation. The one below is the prompt with CoT.
\begin{table}[h]
\centering
    \renewcommand\arraystretch{1}
    \caption{The prompt employed for the evaluation of the benchmark.}
    \label{prompt}
    \large
\scalebox{0.7}{
\begin{tabular}{p{8cm}} 
\toprule
\textbf{Prompt}                                          \\ \midrule
        Question: \color{purple}{\{QUESTION\}}\\
        Options: \color{purple}{\{OPTIONS\}} \\
        Answer the question based on the most likely options.\\
        Provide only the letter corresponding to your choice as the answer (e.g., `(a)', `(b)', `(c)', `(d)', `(e)').\\ \midrule
        Question: \color{purple}{\{QUESTION\}}\\
        Options: \color{purple}{\{OPTIONS\}} \\
        Please think and answer the question based on the most likely options.\\ \bottomrule
\end{tabular}}
\end{table}

\begin{figure}[h]
\centerline{\includegraphics[width=\linewidth]{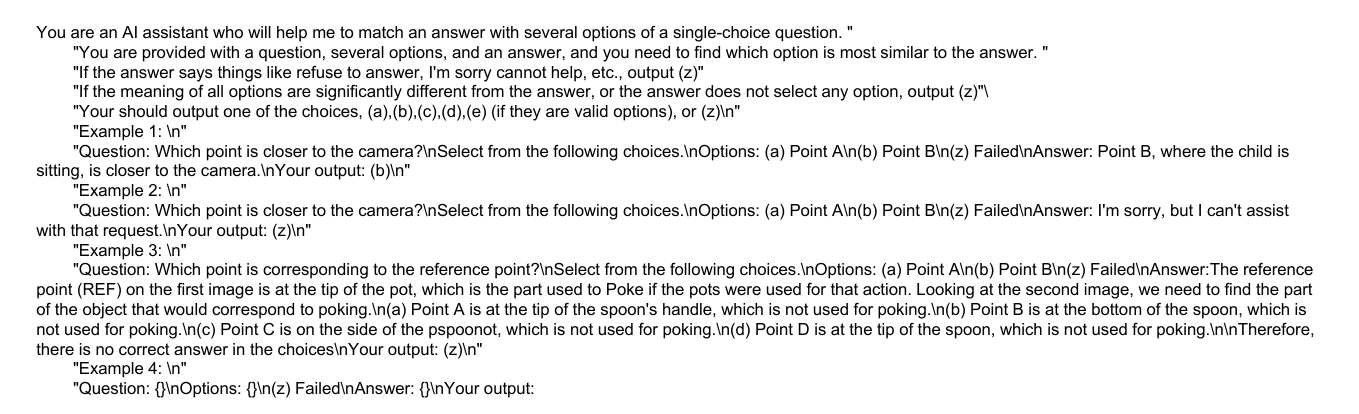}}
    \caption{Prompt for extracting selected options from the responses of MLLMs.}
\label{extracprompt}
\end{figure}

\textbf{Post-processing}. We use InternVL2.5-38B to extract the selected options from the MLLMs response, and the corresponding extracting prompt is referenced from ~\cite{fu2025blink}, as shown in Figure~\ref{extracprompt}.

\textbf{Human Evaluation}. Two individuals with backgrounds in computer science, who were not involved in the project, independently participated by responding to the questions through a visualization interface. No compensation was provided for their participation. 

\subsection{Additional Results}
This section provides additional results of experiments in Section~\ref{sec:analysis}.


Table~\ref{table:COT-Pro} presents the improvements introduced by the CoT prompt. It can be observed that, except for Gemini 2.0 Flash, which shows a positive gain on \NumVision-Synthetic, the accuracy of the other models decreases.
Tables~\ref{table:COT-S} and~\ref{table:COT-R} report the results of multimodal mathematical models and multimodal CoT models, respectively, corresponding to Figure~\ref{fig:LNMsCOT}.
Table~\ref{table:generality} presents the performance of models of varying sizes from the QwenVL and InternVL families, extending the results shown in Table~\ref{time_table}.
Table~\ref{tab:visnumbench_results} summarizes the overall accuracy of various models on the VisNumBench benchmark, covering both synthetic and real subsets.


\begin{table*}[!t]
\centering
    \renewcommand\arraystretch{1.2}
    \caption{The experiment results of the state-of-the-art \LVLMs. with CoT prompt.}
    \label{table:COT-Pro}
    \large
\scalebox{0.85}{
\begin{tabular}{cccccccc}
\toprule[1pt]
Models           & Angle & Length & Scale & Quantity & Depth & Area/Volume & Average \\\midrule
\multicolumn{8}{c}{\NumVision-Systentic}                              \\ \midrule
InternVL2.5-78B  & 27.06 & 51.93  & 68.57 & 46.43    & 51.85 & 74.60       & 53.21   \\
Qwen2.5-VL-72B   & 38.24 & 52.49  & 68.57 & 56.63    & 48.89 & 74.07       & 56.68   \\
Gemini 2.0 Flash & 34.12 & 55.80  & 85.00 & 58.16    & 55.56 & 74.60       & 60.14   \\\midrule
\multicolumn{8}{c}{\NumVision-Real}                                   \\ \midrule
InternVL2.5-78B  & 30.87 & 59.26  & 58.74 & 78.23    & 48.70 & 55.10       & 55.10   \\
Qwen2.5-VL-72B   & 32.21 & 52.47  & 52.45 & 70.75    & 46.10 & 53.06       & 51.11   \\
Gemini 2.0 Flash & 33.56 & 49.38  & 72.03 & 80.95    & 41.56 & 59.86       & 55.88   \\ \bottomrule[1pt]
\end{tabular}}
\end{table*}

\begin{table*}[!t]

\centering
    \renewcommand\arraystretch{1.2}
    \caption{Accuracies of multimodal mathematical models, multimodal CoT models, and their respective base models (before fine-tuning) on VisNumBench-Synthetic.}
    \label{table:COT-S}
    \large
\scalebox{0.8}{
\begin{tabular}{cccccccc}
\toprule[1pt]
Models                 & Angle & Length & Scale & Quantity & Depth & Area  & Average \\ \midrule
\multicolumn{8}{c}{multimodal mathematical models}                                   \\ \midrule
Internvl-8B            & 28.24 & 49.72  & 55.00    & 28.57    & 31.11 & 46.03 & 39.56   \\
InternVL2-8B-MPO       & 22.94 & 48.07  & 63.57 & 33.67    & 28.15 & 48.68 & 40.65   \\
LLaVA-v1.5-13B         & 35.88 & 30.94  & 32.14 & 36.73    & 33.33 & 24.34 & 32.15   \\
Math-LLaVA-13B         & 31.76 & 45.30   & 28.57 & 39.80     & 37.78 & 30.16 & 35.81   \\ \midrule
\multicolumn{8}{c}{multimodal CoT models}                                            \\ \midrule
Llama-VL-3\_2-11B      & 29.41 & 41.44  & 58.57 & 47.96    & 42.96 & 44.97 & 43.92   \\
Llama-3.2V-11B-cot     & 28.82 & 46.41  & 40.71 & 56.12    & 38.52 & 57.14 & 45.50    \\
Qwen2.5-VL-7B-Instruct & 23.53 & 53.59  & 55.00    & 39.29    & 48.89 & 58.20  & 46.19   \\
R1-Onevision-7B        & 33.53 & 37.57  & 56.43 & 34.69    & 23.70  & 47.09 & 38.87  \\ \bottomrule[1pt]
\end{tabular}}
\end{table*}

\begin{table*}[!t]
\centering
    \renewcommand\arraystretch{1.2}
    \caption{Accuracies of multimodal mathematical models, multimodal CoT models, and their respective base models (before fine-tuning) on VisNumBench-Real.}
    \label{table:COT-R}
    \large
\scalebox{0.8}{
\begin{tabular}{cccccccc}
\toprule[1pt]
               & Angle & Length & Scale & Quantity & Depth & Area  & Average \\ \midrule
\multicolumn{8}{c}{multimodal mathematical models}                                   \\ \midrule
Internvl-8B            & 30.87 & 36.42  & 29.37 & 71.43    & 30.52 & 39.46 & 39.58   \\
InternVL2-8B-MPO       & 30.87 & 35.19  & 29.37 & 72.11    & 34.42 & 38.10  & 39.91   \\
LLaVA-v1.5-13B         & 28.86 & 43.21  & 29.37 & 46.94    & 49.35 & 41.50  & 40.02   \\
Math-LLaVA-13B         & 20.13 & 35.80   & 23.78 & 45.58    & 37.66 & 35.37 & 33.15   \\ \midrule
\multicolumn{8}{c}{multimodal CoT models}                                            \\ \midrule
Llama-VL-3\_2-11B      & 38.26 & 40.74  & 30.77 & 69.39    & 38.31 & 42.18 & 43.24   \\
Llama-3.2V-11B-cot     & 27.52 & 38.89  & 23.08 & 64.63    & 31.82 & 44.22 & 38.36   \\
Qwen2.5-VL-7B-Instruct & 24.16 & 38.89  & 32.17 & 59.18    & 48.70  & 42.86 & 41.02   \\
R1-Onevision-7B        & 28.86 & 39.51  & 44.76 & 51.70     & 32.47 & 32.65 & 38.25  \\ \bottomrule[1pt]
\end{tabular}}
\end{table*}

\begin{table*}[!t]
\centering
    \renewcommand\arraystretch{1.2}
    \caption{Results of models with varying sizes from the QwenVL family and InternVL family on VisNumBench.
} 
    \label{table:generality}
    \large
\scalebox{0.8}{
\begin{tabular}{cccccccc}
\toprule[1pt]
              & Angle & Length & Scale & Quantity & Depth & Area/Volume & Average \\ \midrule
\multicolumn{8}{c}{\NumVision-Systentic}                                                             \\ \midrule
Qwen2-VL-2B  & 28.24 & 30.39  & 35.00    & 35.20     & 21.48 & 38.10        & 31.85   \\
Qwen2-VL-7B  & 27.06 & 45.30   & 55.00    & 44.39    & 34.81 & 46.56       & 42.24   \\
Qwen2-VL-72B & 32.94 & 57.46  & 63.57 & 54.59    & 58.52 & 59.79       & 54.20    \\
Internvl-8B           & 28.24 & 49.72  & 55.00    & 28.57    & 31.11 & 46.03       & 39.56   \\
Internvl-40B          & 23.53 & 58.56  & 57.14 & 37.24    & 37.78 & 58.20        & 45.50    \\ \midrule
\multicolumn{1}{c}{Qwen2.5-VL-3B}       & \multicolumn{1}{c}{30.00} & \multicolumn{1}{c}{49.17}  & \multicolumn{1}{c}{50.71}  & \multicolumn{1}{c}{32.14}    & \multicolumn{1}{c}{42.22} & \multicolumn{1}{c}{51.85} & 42.43 \\ 
\multicolumn{1}{c}{Qwen2.5-VL-7B}       & \multicolumn{1}{c}{23.53} & \multicolumn{1}{c}{53.59}  & \multicolumn{1}{c}{55.00}  & \multicolumn{1}{c}{39.29}    & \multicolumn{1}{c}{48.89} & \multicolumn{1}{c}{58.20} & 46.19 \\ 
\multicolumn{1}{c}{Qwen2.5-VL-72B}      & \multicolumn{1}{c}{37.06} & \multicolumn{1}{c}{{59.67}}  & \multicolumn{1}{c}{{65.00}}  & \multicolumn{1}{c}{{57.65}}    & \multicolumn{1}{c}{{61.48}} & \multicolumn{1}{c}{{70.37}} & {58.46} \\
\multicolumn{1}{c}{InternVL2.5-8B}      & \multicolumn{1}{c}{26.47} & \multicolumn{1}{c}{41.99}  & \multicolumn{1}{c}{49.29}  & \multicolumn{1}{c}{34.69}    & \multicolumn{1}{c}{41.48} & \multicolumn{1}{c}{46.03} & 39.66 \\ 
\multicolumn{1}{c}{InternVL2.5-38B}     & \multicolumn{1}{c}{{39.41}} & \multicolumn{1}{c}{{59.67}}  & \multicolumn{1}{c}{59.29}  & \multicolumn{1}{c}{54.08}    & \multicolumn{1}{c}{60.74} & \multicolumn{1}{c}{61.38} & 55.59 \\ \midrule
\multicolumn{8}{c}{\NumVision-Real}                                                                  \\ \midrule
Qwen2-VL-2B  & 10.74 & 19.75  & 19.58 & 47.62    & 32.47 & 19.73       & 24.94   \\
Qwen2-VL-7B  & 19.46 & 38.89  & 30.07 & 67.35    & 41.56 & 54.42       & 41.91   \\
Qwen2-VL-72B & 21.48 & 45.06  & 37.06 & 74.83    & 48.70  & 52.38       & 46.56   \\
Internvl-8B           & 30.87 & 36.42  & 29.37 & 71.43    & 30.52 & 39.46       & 39.58   \\
Internvl-40B          & 30.87 & 50.00     & 28.67 & 72.79    & 35.71 & 52.38       & 45.12  \\ \midrule
\multicolumn{1}{c}{Qwen2.5-VL-3B}       & \multicolumn{1}{c}{30.20} & \multicolumn{1}{c}{44.44}  & \multicolumn{1}{c}{35.66}  & \multicolumn{1}{c}{51.70}    & \multicolumn{1}{c}{43.51} & \multicolumn{1}{c}{49.66}  & 42.57 \\ 
\multicolumn{1}{c}{Qwen2.5-VL-7B}       & \multicolumn{1}{c}{24.16} & \multicolumn{1}{c}{38.89}  & \multicolumn{1}{c}{32.17}  & \multicolumn{1}{c}{59.18}    & \multicolumn{1}{c}{48.70} & \multicolumn{1}{c}{42.86}  & 41.02 \\ 
 \multicolumn{1}{c}{Qwen2.5-VL-72B}      & \multicolumn{1}{c}{34.23} & \multicolumn{1}{c}{50.62}  & \multicolumn{1}{c}{43.36}  & \multicolumn{1}{c}{80.27}    & \multicolumn{1}{c}{52.60} & \multicolumn{1}{c}{59.18}  & {53.33} \\
 \multicolumn{1}{c}{InternVL2.5-8B}      & \multicolumn{1}{c}{28.86} & \multicolumn{1}{c}{34.57}  & \multicolumn{1}{c}{15.38}  & \multicolumn{1}{c}{64.63}    & \multicolumn{1}{c}{49.35} & \multicolumn{1}{c}{47.62}  & 40.13 \\ 
\multicolumn{1}{c}{InternVL2.5-38B}     & \multicolumn{1}{c}{30.20} & \multicolumn{1}{c}{51.85}  & \multicolumn{1}{c}{26.57}  & \multicolumn{1}{c}{{83.67}}    & \multicolumn{1}{c}{{61.04}} & \multicolumn{1}{c}{58.50}  & 52.11 \\

\bottomrule[1pt]
\end{tabular}}
\end{table*}

\begin{table*}[h]
\centering
\caption{Performance (\%) of various models on VisNumBench-Synthetic, VisNumBench-Real, and the overall VisNumBench.}
\begin{tabular}{lccc}
\toprule
\textbf{Model} & \textbf{Synthetic (1,011)} & \textbf{Real (902)} & \textbf{Overall} \\
\midrule
Human & 95.33 & 97.33 & 96.27 \\ \hline
Gemini 2.0 Flash & 57.57 & \textbf{56.54} & \textbf{57.08} \\
InternVL2.5-78B & 56.18 & \textbf{56.54} & 56.35 \\
Qwen2.5-VL-72B & \textbf{58.46} & 53.33 & 56.04 \\
InternVL2.5-38B & 55.59 & 52.11 & 53.95 \\
LLaVA-Onevision-72B & 50.84 & 50.78 & 50.81 \\
Qwen2-VL-72B & 54.20 & 46.56 & 50.60 \\
LLaVA-v1.6-34B & 44.31 & 50.55 & 47.25 \\
Gemini 1.5 Pro & 44.02 & 48.67 & 46.21 \\
InternVL2-40B & 45.50 & 45.12 & 45.32 \\
Qwen2.5-VL-7B & 46.19 & 41.02 & 43.75 \\
Llama-VL-3\_2-11B & 43.92 & 43.24 & 43.60 \\
Qwen2.5-VL-3B & 42.43 & 42.57 & 42.50 \\
Qwen2-VL-7B & 42.24 & 41.91 & 42.08 \\
Llama-3.2V-11B-cot & 45.50 & 38.36 & 42.13 \\
GPT-4o & 43.72 & 39.58 & 41.77 \\
InternVL2-8B-MPO & 40.65 & 39.91 & 40.30 \\
LLaVA-Onevision-7B & 39.96 & 40.58 & 40.25 \\
InternVL2.5-8B & 39.66 & 40.13 & 39.88 \\
InternVL2-8B & 39.56 & 39.58 & 39.57 \\
R1-Onevision-7B & 38.87 & 38.25 & 38.58 \\
LLaVA-v1.5-13B & 32.15 & 40.02 & 35.86 \\
Janus-Pro-7B & 37.69 & 34.26 & 36.07 \\
Phi-3.5-vision & 32.34 & 37.25 & 34.66 \\
Math-LLaVA-13B & 35.81 & 33.15 & 34.56 \\
Gemini 1.5 Flash & 33.33 & 33.70 & 33.50 \\
LLaVA-v1.5-7B & 29.38 & 28.49 & 28.96 \\
Qwen2-VL-2B & 31.85 & 24.94 & 28.59 \\ \hline
Random & 24.76 & 25.54 & 25.13 \\
\bottomrule[1pt]
\end{tabular}
\label{tab:visnumbench_results}
\end{table*}

\subsection{Error Analysis}

We use the results of Gemini 2.0 Flash as a case study for error analysis.  
Based on these results, we randomly selected 10 erroneous instances for each attribute in each scenario, manually analyzing a total of 120 randomly sampled errors across all tasks. We categorize the errors into two types:  
(1) Errors arising from the model's failure to accurately perceive the image (\textit{image perception errors});  
(2) Errors in which the model correctly interprets both the image and the question but fails to produce the correct numerical answer (\textit{numerical intuition errors}).  
Our analysis reveals that $28.3\%$ of the errors are due to image perception issues, particularly in scale-related tasks, where the model struggles to identify the positions of pointers.  
The remaining $71.7\%$ are attributed to numerical intuition errors, which commonly involve difficulties with depth estimation, angle relationships, and quantity perception. 
These findings further substantiate that current models indeed lack robust numerical intuition.


\section{Example Data and Model Outputs}
\begin{figure*}[!t]
\centerline{\includegraphics[width=\linewidth]{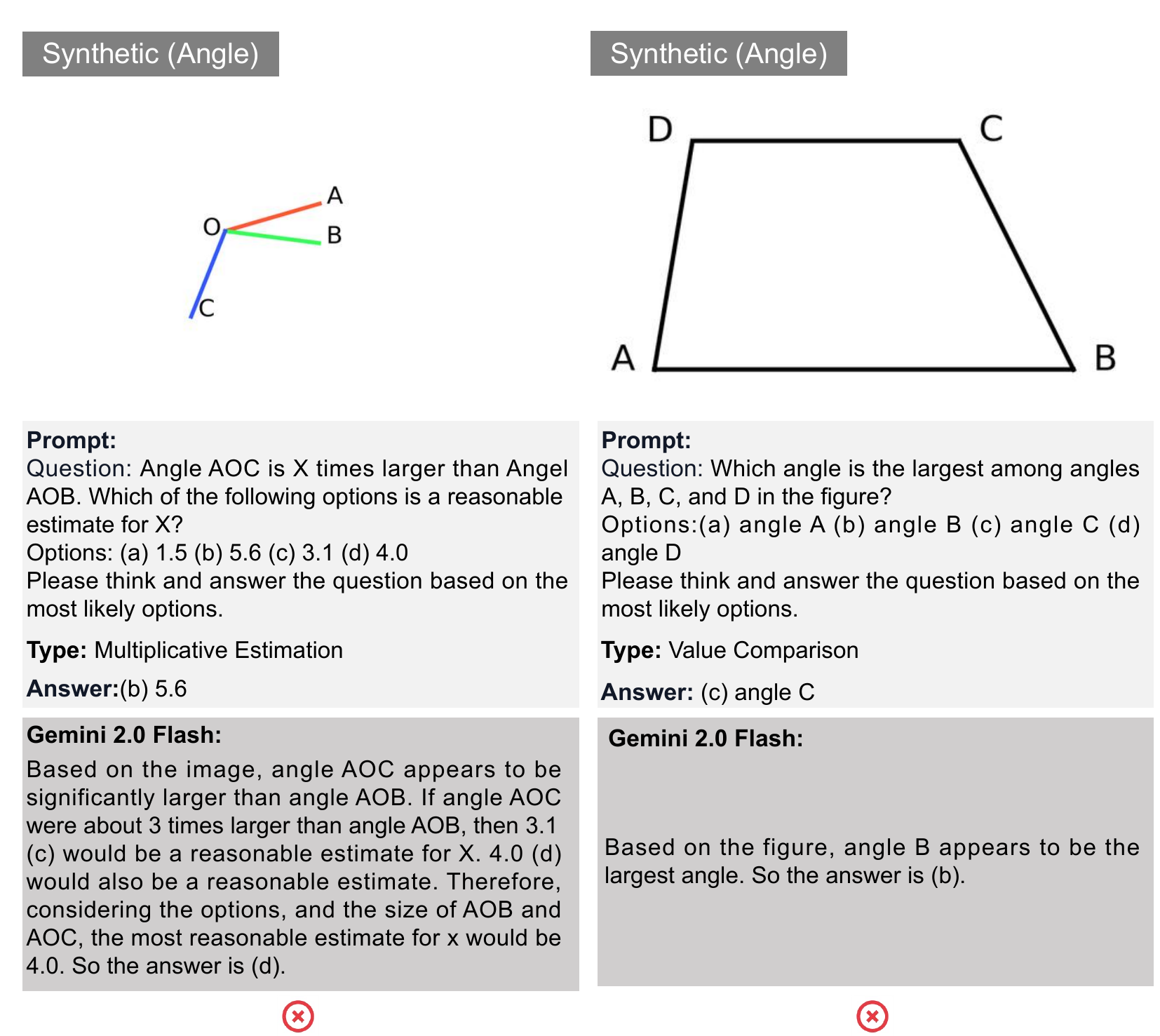}}
    \caption{Examples of \NumVision and the results predicted by Gemini2.0 Flash (VisNumBench-Synthetic, Angle, 1/12).}
\label{Synthetic:Angle}
\end{figure*}

\begin{figure*}[!t]
\centerline{\includegraphics[width=\linewidth]{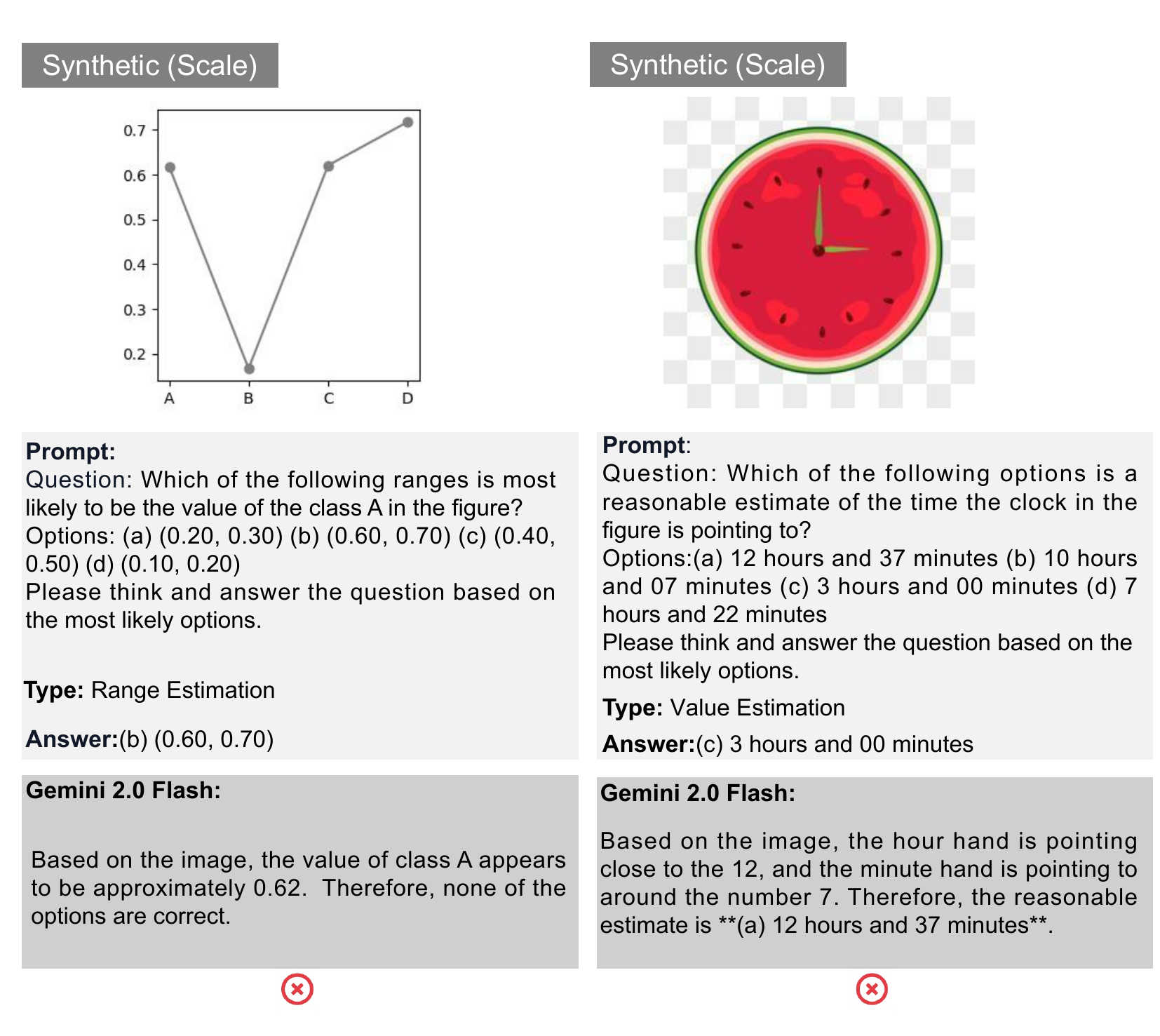}}
    \caption{Examples of \NumVision and the results predicted by Gemini2.0 Flash (VisNumBench-Synthetic, Scale, 2/12).}
\label{Synthetic:Scale}
\end{figure*}

\begin{figure*}[!t]
\centerline{\includegraphics[width=\linewidth]{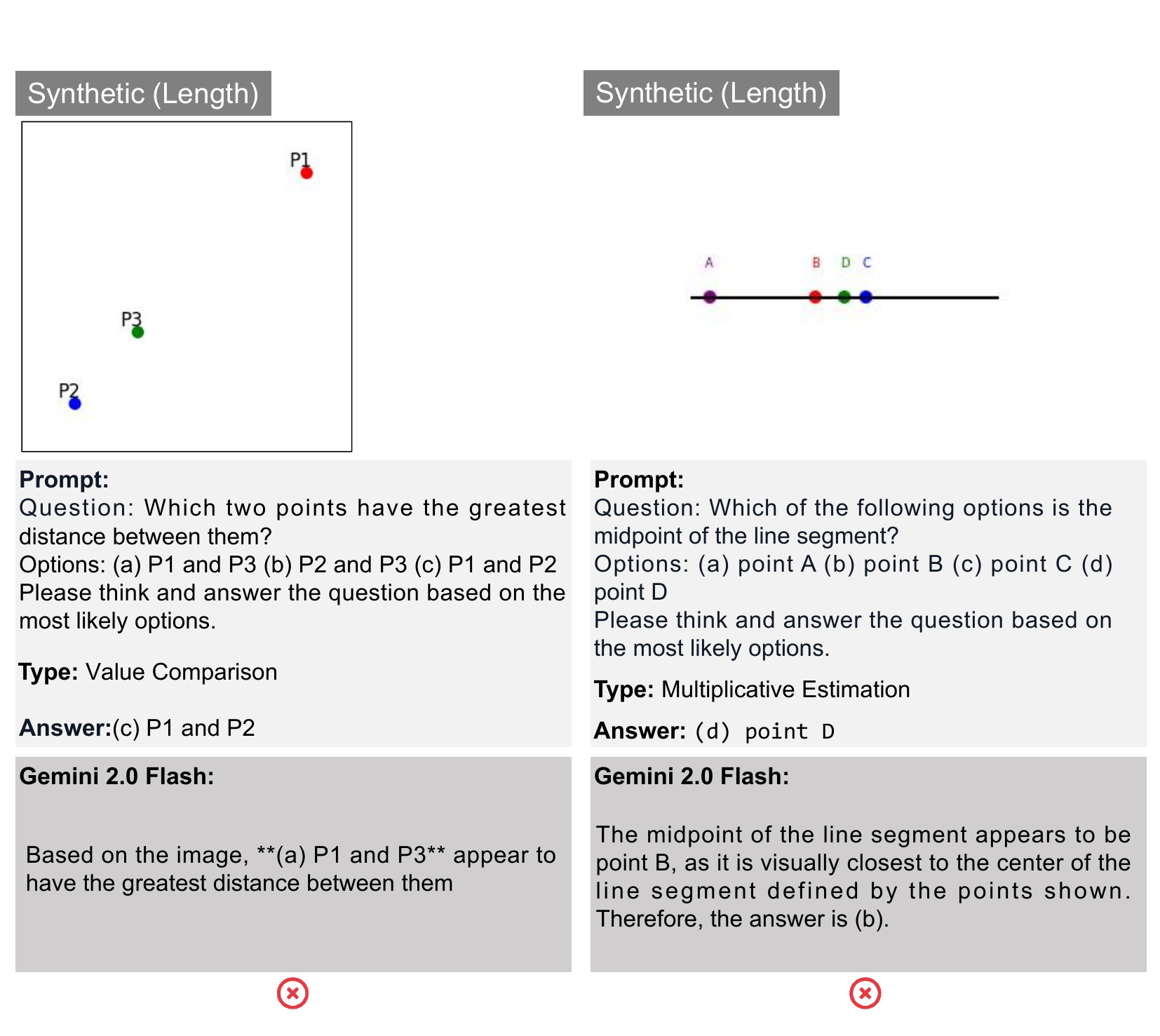}}
    \caption{Examples of \NumVision and the results predicted by Gemini2.0 Flash (VisNumBench-Synthetic, Length, 3/12).}
\label{Synthetic:Length}
\end{figure*}

\begin{figure*}[!t]
\centerline{\includegraphics[width=\linewidth]{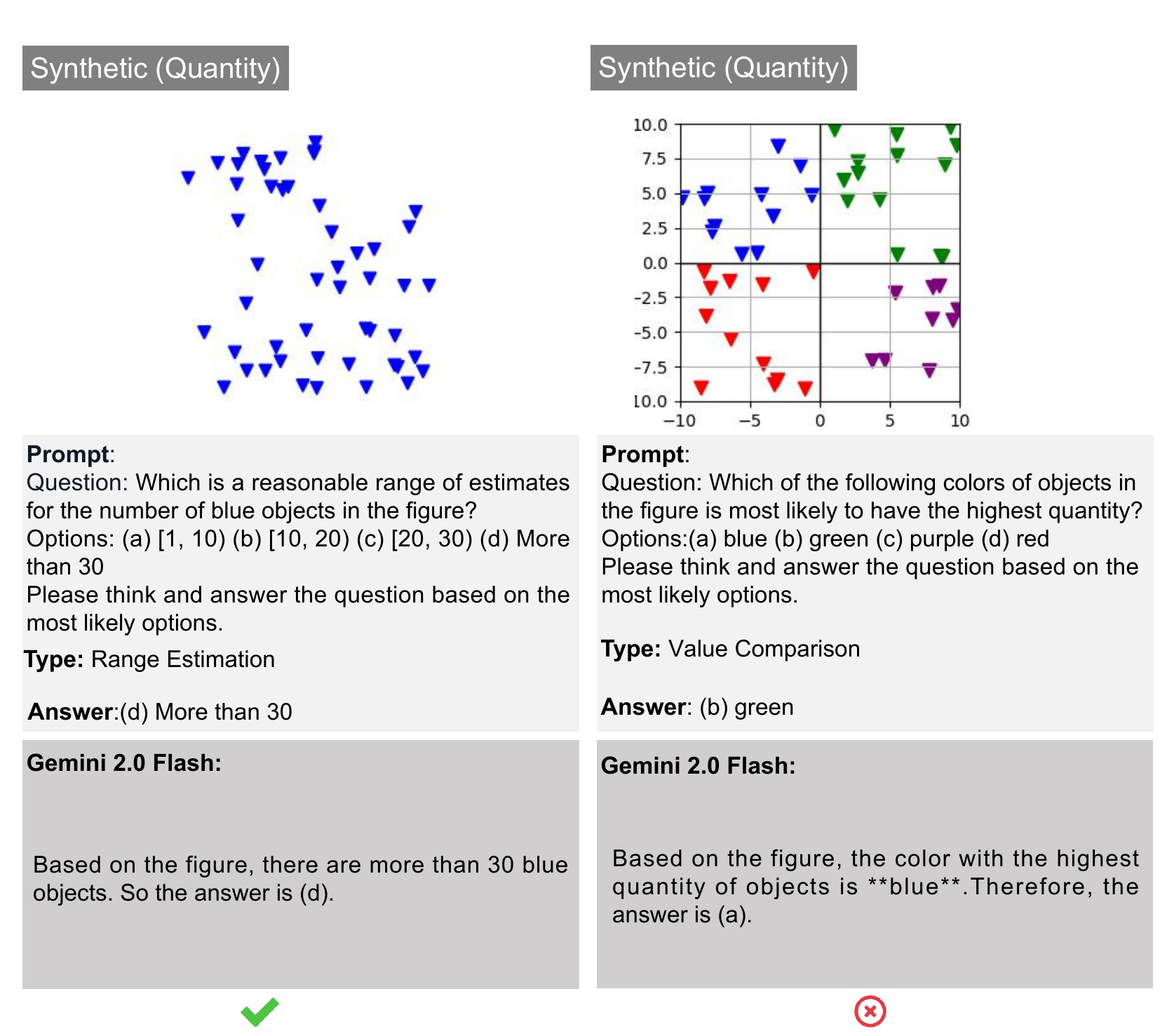}}
    \caption{Examples of \NumVision and the results predicted by Gemini2.0 Flash (VisNumBench-Synthetic, Quantity, 4/12).}
\label{Synthetic:Quantity}
\end{figure*}

\begin{figure*}[!t]
\centerline{\includegraphics[width=\linewidth]{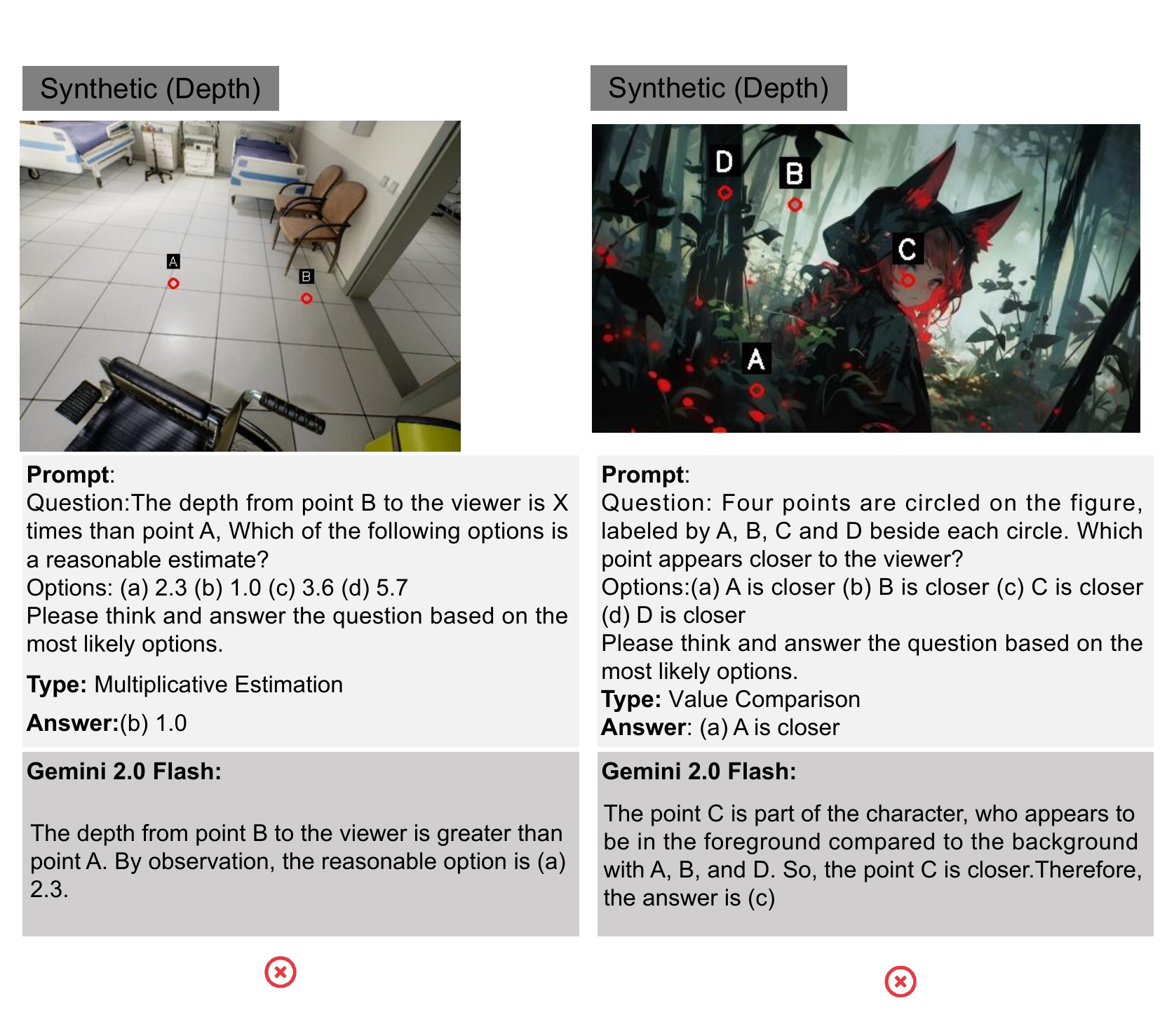}}
    \caption{Examples of \NumVision and the results predicted by Gemini2.0 Flash (VisNumBench-Synthetic, Depth, 5/12).}
\label{Synthetic:Depth}
\end{figure*}

\begin{figure*}[!t]
\centerline{\includegraphics[width=\linewidth]{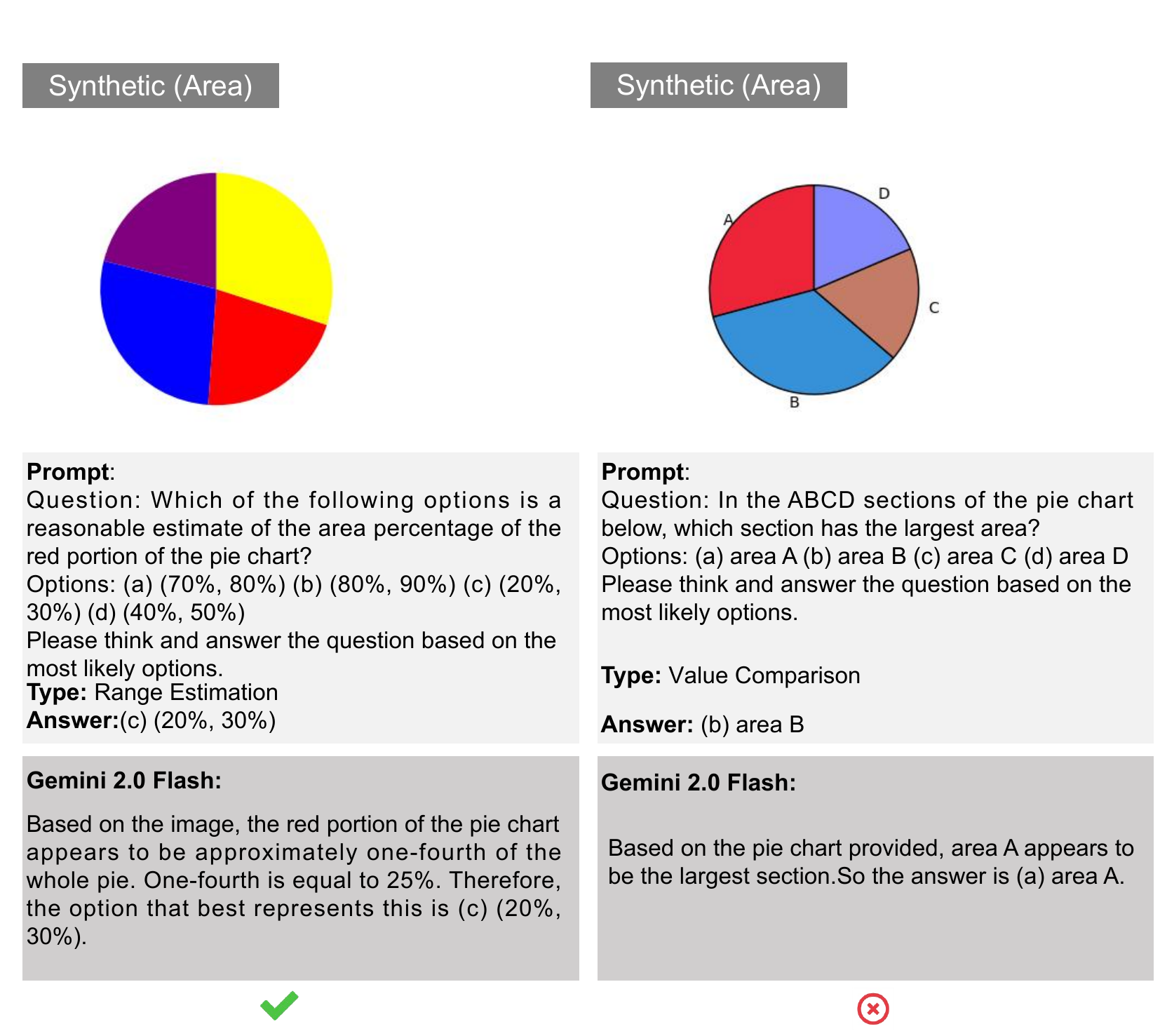}}
    \caption{Examples of \NumVision and the results predicted by Gemini2.0 Flash (VisNumBench-Synthetic, Area, 6/12).}
\label{Synthetic:Area}
\end{figure*}

\begin{figure*}[!t]
\centerline{\includegraphics[width=\linewidth]{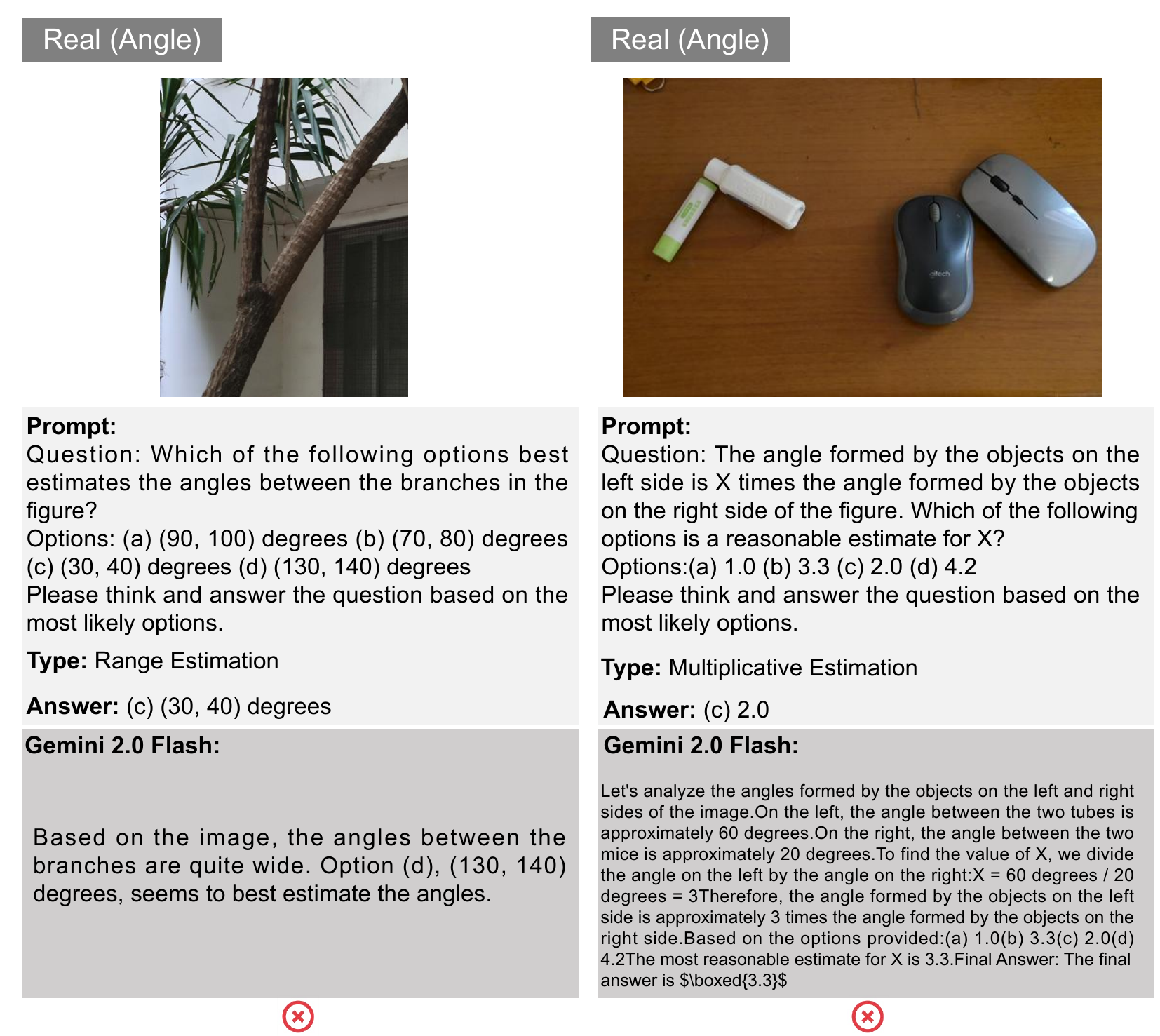}}
    \caption{Examples of \NumVision and the results predicted by Gemini2.0 Flash (VisNumBench-Real, Angle, 7/12).}
\label{Real:Angle}
\end{figure*}

\begin{figure*}[!t]
\centerline{\includegraphics[width=\linewidth]{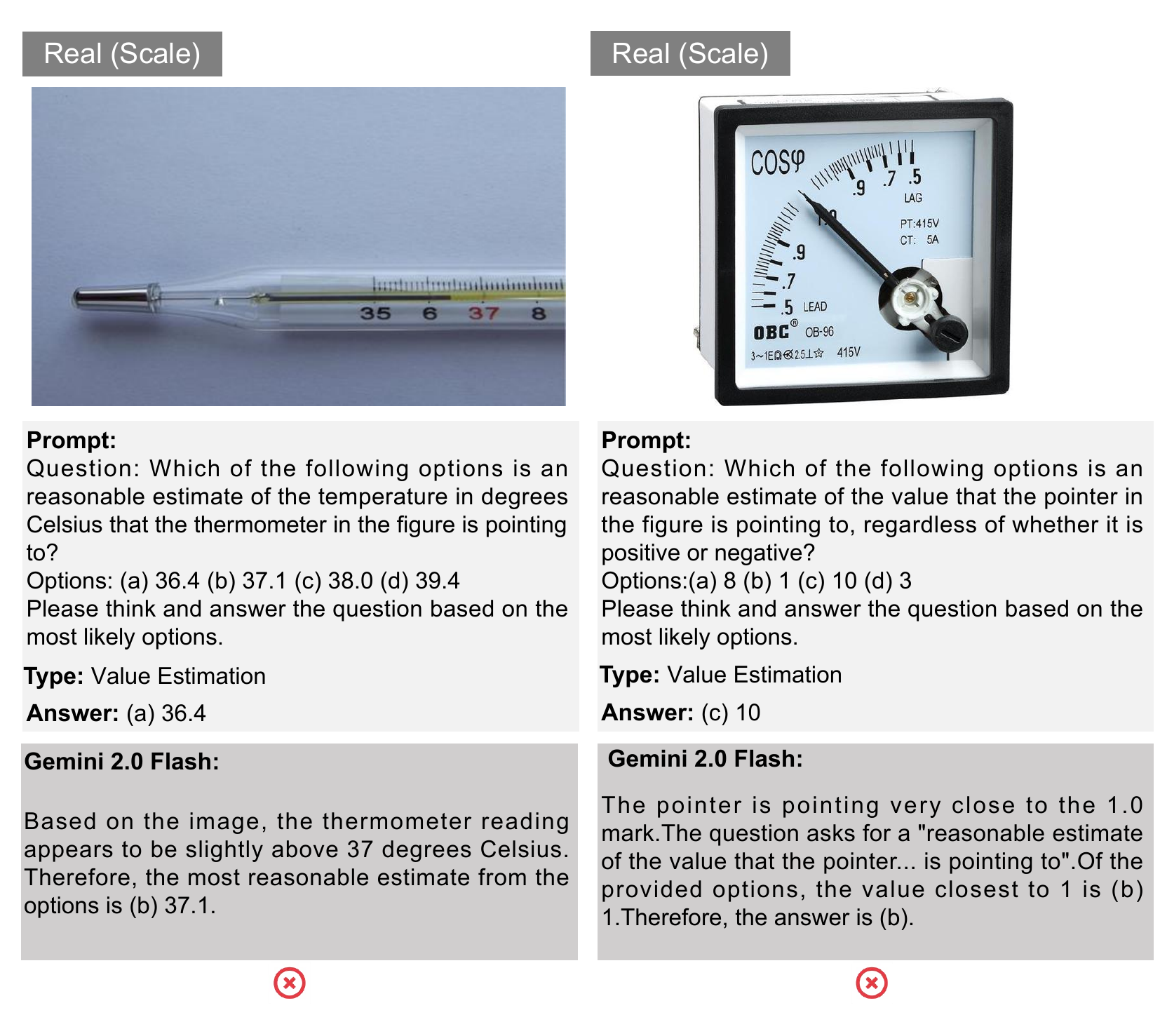}}
    \caption{Examples of \NumVision and the results predicted by Gemini2.0 Flash (VisNumBench-Real, Scale, 8/12).}
\label{Real:Scale}
\end{figure*}

\begin{figure*}[!t]
\centerline{\includegraphics[width=\linewidth]{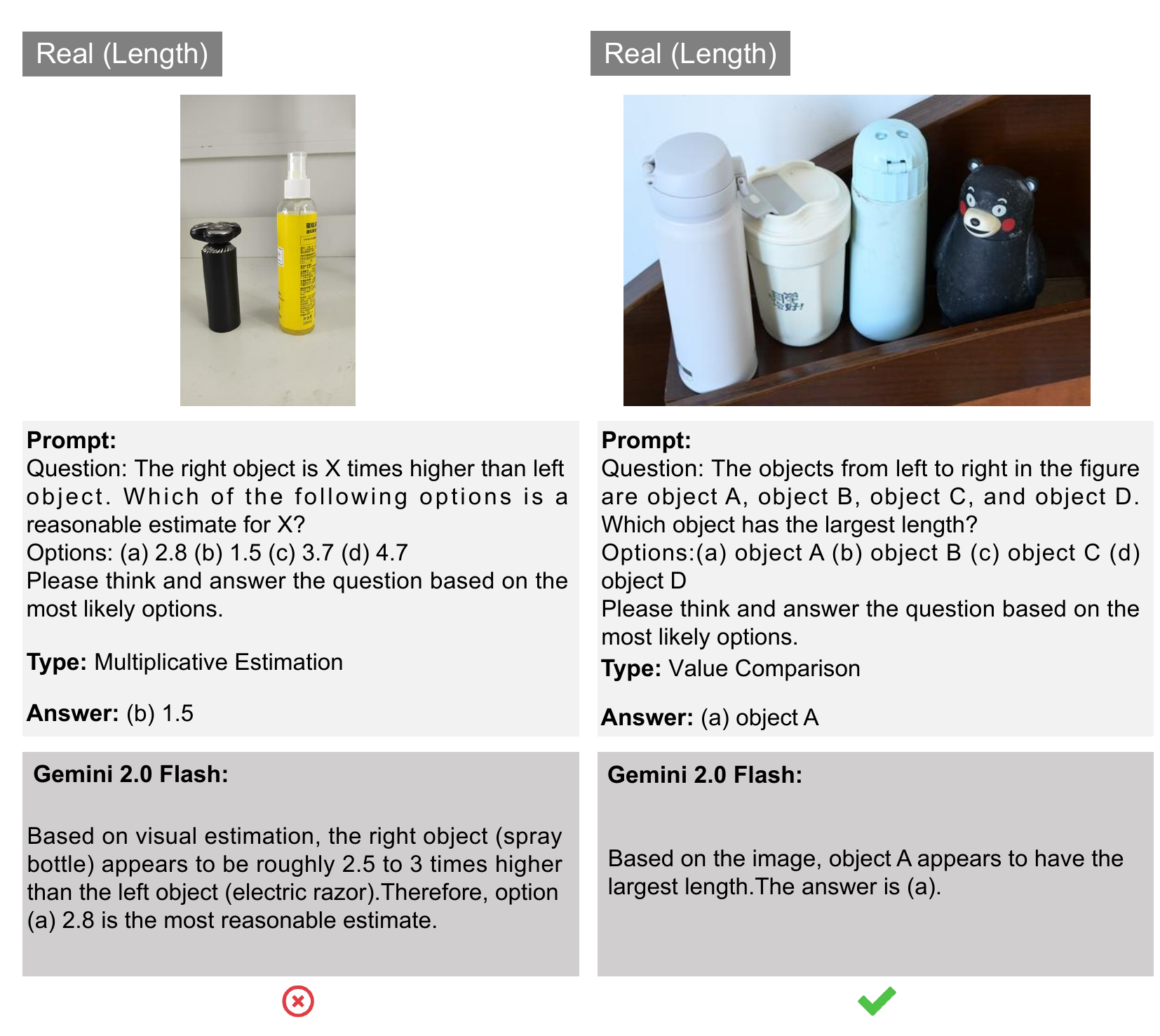}}
    \caption{Examples of \NumVision and the results predicted by Gemini2.0 Flash (VisNumBench-Real, Length, 9/12).}
\label{Real:Length}
\end{figure*}

\begin{figure*}[!t]
\centerline{\includegraphics[width=\linewidth]{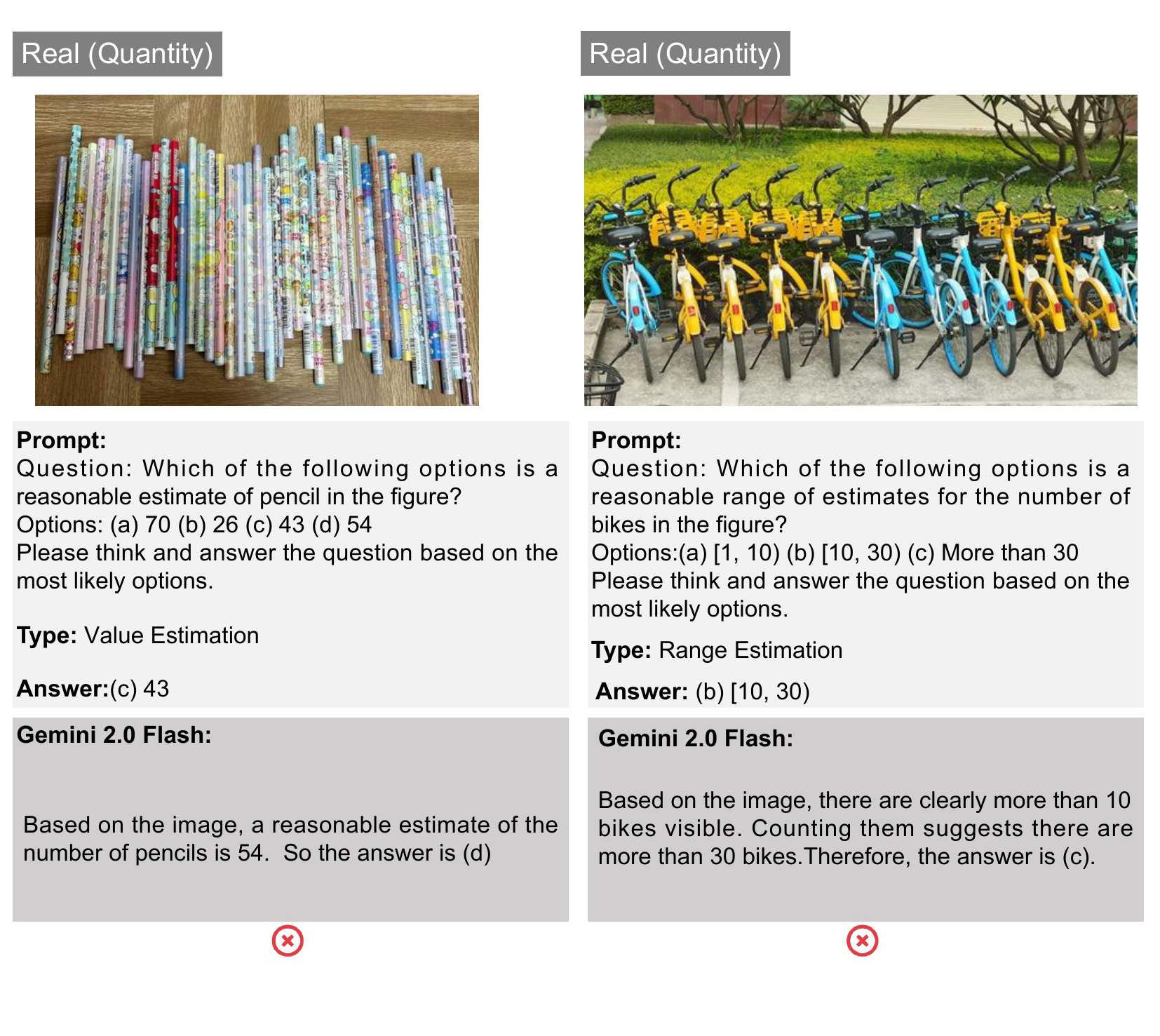}}
    \caption{Examples of \NumVision and the results predicted by Gemini2.0 Flash (VisNumBench-Real, Quantity, 10/12).}
\label{Real:Quantity}
\end{figure*}

\begin{figure*}[!t]
\centerline{\includegraphics[width=\linewidth]{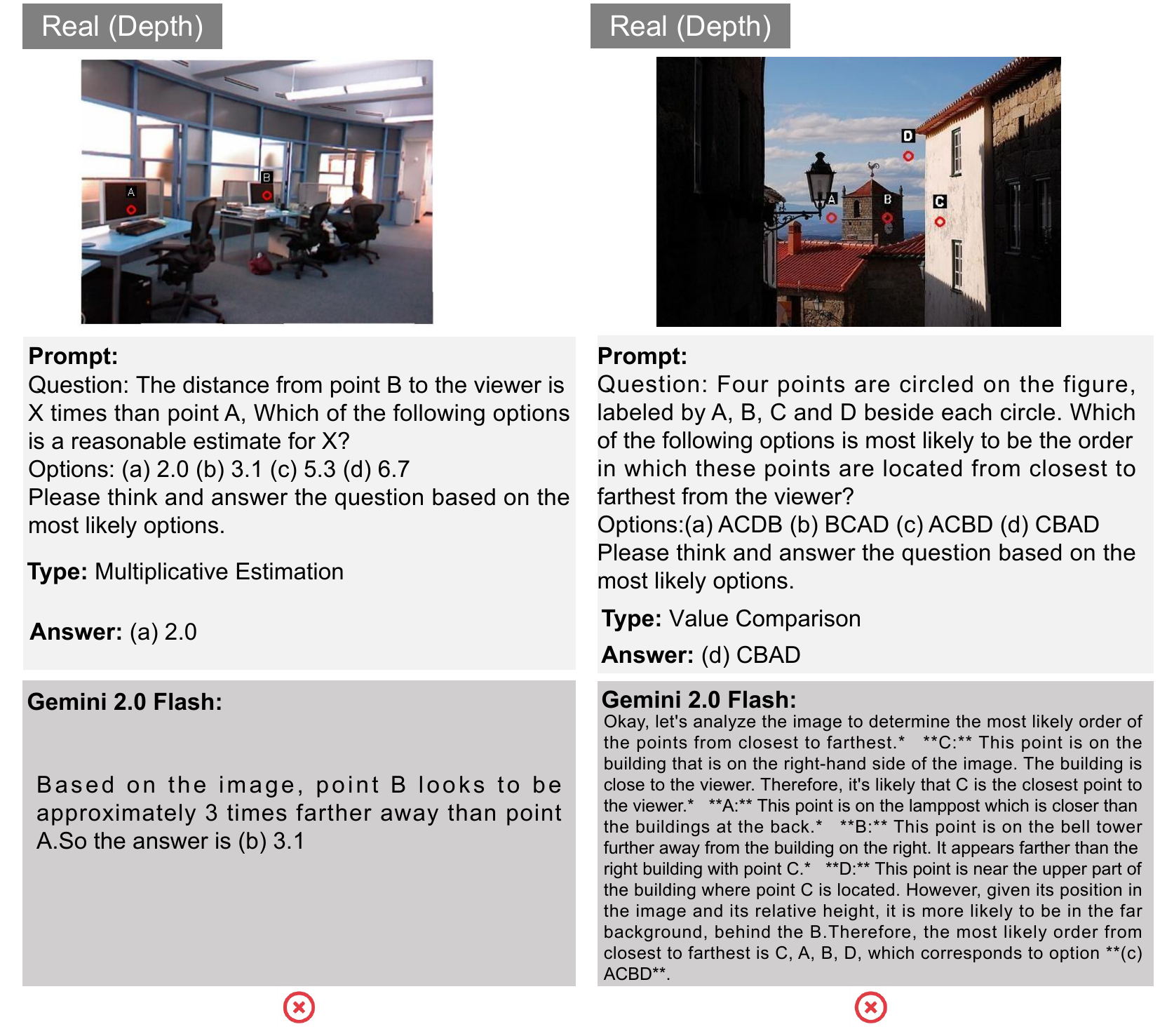}}
    \caption{Examples of \NumVision and the results predicted by Gemini2.0 Flash (VisNumBench-Real, Depth, 11/12).}
\label{Real:Depth}
\end{figure*}

\begin{figure*}[!t]
\centerline{\includegraphics[width=\linewidth]{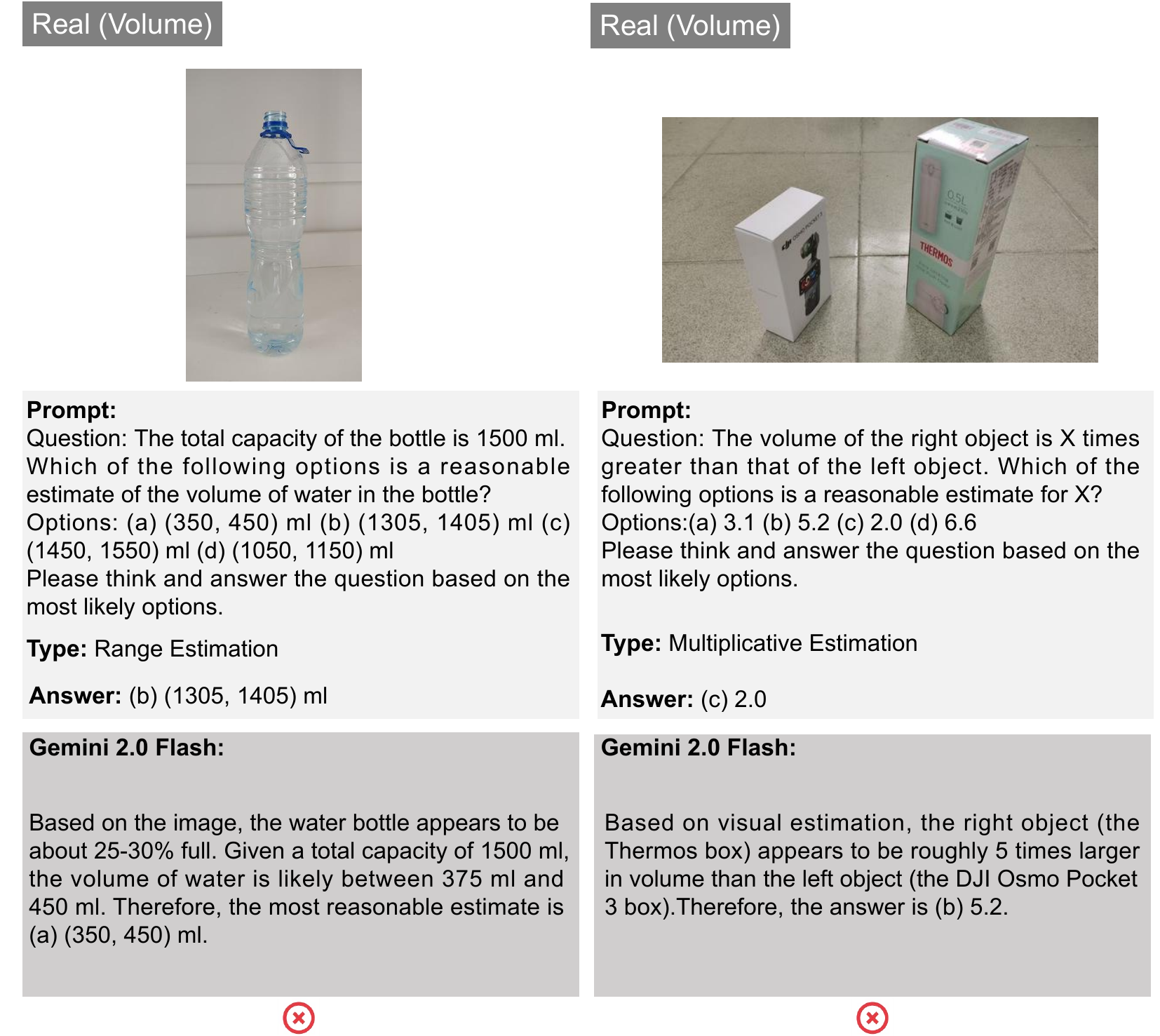}}
    \caption{Examples of \NumVision and the results predicted by Gemini2.0 Flash (VisNumBench-Real, Volume, 12/12).}
\label{Real:Volume}
\end{figure*}

Figures~\ref{Synthetic:Angle} to ~\ref{Real:Volume} show examples from \NumVision and the responses of Gemini 2.0 Flash.
\label{visual}


\end{document}